\documentclass[final,journal,letterpaper,twoside,twocolumn]{IEEEtran}

\usepackage{cite}

\ifCLASSINFOpdf
  \usepackage[pdftex]{graphicx}
  \graphicspath{{./figures/}}
  \DeclareGraphicsExtensions{.eps,.pdf,.jpeg,.png,.tif}
\else
   \usepackage[dvips]{graphicx}
\fi
\usepackage{amsmath, amssymb}
\usepackage{algorithm,algorithmic}
\usepackage{booktabs}
\usepackage{xcolor}
\usepackage{array}

\ifCLASSOPTIONcompsoc
  \usepackage[caption=false,font=normalsize,labelfont=sf,textfont=sf]{subfig}
\else
  \usepackage[caption=false,font=footnotesize]{subfig}
\fi

\begin{document}

\title{Non-Local Compressive Sensing Based SAR Tomography}

\author{Yilei~Shi,~\IEEEmembership{Member,~IEEE},
        Xiao Xiang~Zhu,~\IEEEmembership{Senior~Member,~IEEE},
        and~Richard~Bamler,~\IEEEmembership{Fellow,~IEEE}
\thanks{This work is supported by the European Research Council (ERC) under the European Union's Horizon 2020 research and innovation programme (grant agreement no. ERC-2016-StG-714087, acronym: So2Sat, www.so2sat.eu), the Helmholtz Association under the framework of the Young Investigators Group ``SiPEO" (VH-NG-1018, www.sipeo.bgu.tum.de), Munich Aerospace e.V. – Fakult{\"a}t f{\"u}r Luft- und Raumfahrt, and the Bavaria California Technology Center (Project: Large-Scale Problems in Earth Observation). The authors thank the Gauss Centre for Supercomputing (GCS) e.V. for funding this project by providing computing time on the GCS Supercomputer SuperMUC at the Leibniz Supercomputing Centre (LRZ).}
\thanks{Y.~Shi is with the Chair of Remote Sensing Technology (LMF), Technische Universit{\"a}t M{\"u}nchen (TUM), 80333 Munich, Germany (e-mail: yilei.shi@tum.de)}
\thanks{X.X.~Zhu is with the Remote Sensing Technology Institute (IMF), German Aerospace Center (DLR), and Signal Processing in Earth Observation (SIPEO), Technische Universit{\"a}t M{\"u}nchen (TUM), 80333 Munich, Germany (e-mail: xiaoxiang.zhu@dlr.de)}
\thanks{R. Bamler is with the Remote Sensing Technology Institute (IMF), German Aerospace Center (DLR), and the Chair of Remote Sensing Technology (LMF), Technische Universit{\"a}t M{\"u}nchen (TUM), 80333 Munich, Germany (e-mail: richard.bamler@dlr.de)}
\thanks{\emph{(Correspondence: Xiao Xiang Zhu)}}}

\markboth{submitted to IEEE TRANSACTIONS ON GEOSCIENCE AND REMOTE SENSING, 2018}%
{Y. Shi \MakeLowercase{\textit{et al.}}: Non-Local Compressive Sensing based SAR Tomography}

\maketitle

\begin{abstract}
\textcolor{blue}{This is the preprint version, to read the final version please go to IEEE Transactions on Geoscience and Remote Sensing  on  IEEE  Xplore.}
Tomographic SAR (TomoSAR) inversion of urban areas is an inherently sparse reconstruction problem and, hence, can be solved using compressive sensing (CS) algorithms. This paper proposes solutions for two notorious problems in this field:
1) TomoSAR requires a high number of data sets, which makes the technique expensive. However, it can be shown that the number of acquisitions and the signal-to-noise ratio (SNR) can be traded off against each other, because it is asymptotically only the product of the number of acquisitions and SNR that determines the reconstruction quality. We propose to increase SNR by integrating non-local estimation into the inversion and show that a reasonable reconstruction of buildings from only seven interferograms is feasible.
2) CS-based inversion is computationally expensive and therefore barely suitable for large-scale applications. We introduce a new fast and accurate algorithm for solving the non-local L1-L2-minimization problem, central to CS-based reconstruction algorithms.
The applicability of the algorithm is demonstrated using simulated data and TerraSAR-X high resolution spotlight images over an area in Munich, Germany.
\end{abstract}


\section{Introduction}
\IEEEPARstart{S}{ynthetic} Aperture Radar Tomography (TomoSAR) is an advanced SAR interferometric technique that can not only retrieve 3-D spatial information but also assess the 4-D temporal information, e.g. deformation, in millimeter scale, of individual buildings from meter-resolution SAR satellite data. Repeat-pass multi-baseline SAR tomography has been intensively developed in the past two decades \cite{bib:reigber2000first} \cite{bib:gini2002layover} \cite{bib:lombardini2005differential} \cite{bib:Fornaro2005} \cite{bib:Fornaro2009} \cite{bib:Zhu2010b} \cite{bib:ge2018spaceborne} \cite{bib:zhu2018review} and shows promising results on 3-D reconstruction of urban areas. However, for urban monitoring, there are still several issues that need to be solved: improving the elevation resolution, i.e., providing super-resolution (SR) for layover separation; achieving high 3-D localization accuracy even in the presence of unmodeled, non-Gaussian noise; and retrieving nonlinear motion, e.g., due to seasonal thermal dilation. Driven by these requirements, new algorithms have been invented in the past few years that take advantage of recent developments in signal processing, such as sparse reconstruction and compressive sensing (CS), \cite{bib:zhu2009very} \cite{bib:Zhu2010a} \cite{bib:Budillon2011} \cite{bib:aguilera2013wavelet} \cite{bib:Ma2015}  and can provide height estimates with unprecedented accuracy compared to the state-of-the-art multibaseline InSAR algorithms \cite{bib:Zhu2012c}.

However, CS-based TomoSAR still suffers from two problems for practical use. First, a large number of images are required, typically a stack of 20-100 images over the illuminated area. For instance, it is demonstrated in \cite{bib:Zhu2012b} that by using even the most efficient algorithms, like non-linear least squares (NLS) and SL1MMER, a minimum number of 11 acquisitions is required to achieve a reasonable reconstruction in the interesting parameter range of spaceborne SAR. In \cite{bib:Zhu2015}, a joint sparsity concept was applied to obtain precise TomoSAR reconstruction with only six images by incorporating building a priori knowledge to the estimation. However, due to its demand on precise geometric prior, this method can be only used to reconstruct buildings where the geographic information system (GIS) data is available. The second practical drawback of CS-based TomoSAR is that it suffers from a high computational expense and is hard to extend to large-scale practice. Wang et. al. \cite{bib:Wang2014} proposed an efficient approach to address this issue, which uses the well-established and computationally efficient persistent scatterer interferometry to obtain a priori knowledge of the estimates, followed by the linear method and the CS-based SL1MMER algorithm applied to different pre-classified groups of pixels. This approach speeds up the processing, but only to the extent that it reduces the percentage of pixels that require sparse reconstruction.

In this work, we propose a novel framework for TomoSAR with a minimum number of acquisitions to obtain a fast and accurate estimation of elevation without any a priori knowledge. It is mainly motivated by the recent advances in non-local means approaches \cite{bib:buades2005non}\cite{bib:dabov2007image} in image restoration. Non-local means approaches successfully achieve state-of-the-art performance in image restoration \cite{bib:dabov2007image} by seeking the correlation of image patches. As a common prior in natural images, the patch correlation should help increase the signal-to-noise ratio (SNR) of the original signal. As comprehensively investigated in \cite{bib:Zhu2012b}, the product of the number of acquisitions and SNR determines the reconstruction quality, which means that an increase of SNR can dramatically reduce the number of acquisitions needed for reconstruction.

The main contributions of this paper are listed as follows.
\begin{itemize}
\item We propose a novel framework, ``non-local compressive sensing (NLCS) TomoSAR'',  to produce accurate estimates of scatterers' position without any a priori knowledge, using as few images as possible.
\item We further propose an efficient algorithm to solve the NLCS model, containing an optimized parallelization scheme for a non-local process and a fast and accurate solver for complex-valued $L_1$ least squares minimization.
\item Systematic performance evaluation of the proposed approach has been carried out using both simulated and real data. The results show that the proposed method can produce very accurate estimations of elevation without notable resolution distortion.
\end{itemize}

The paper is organized as follows. In section II, the SAR imaging model and TomoSAR inversion with compressive sensing approach are introduced. In section III, a novel approach called ``non-local compressive sensing based TomoSAR'' is introduced. Experiments using simulated data and real data are presented in section IV and VI. Finally, conclusions are given in section VII.

\section{CS-based SAR Tomography}
\subsection{SAR Imaging Model}
\begin{figure}[h!]
  \centering
  \includegraphics[width=0.5\textwidth]{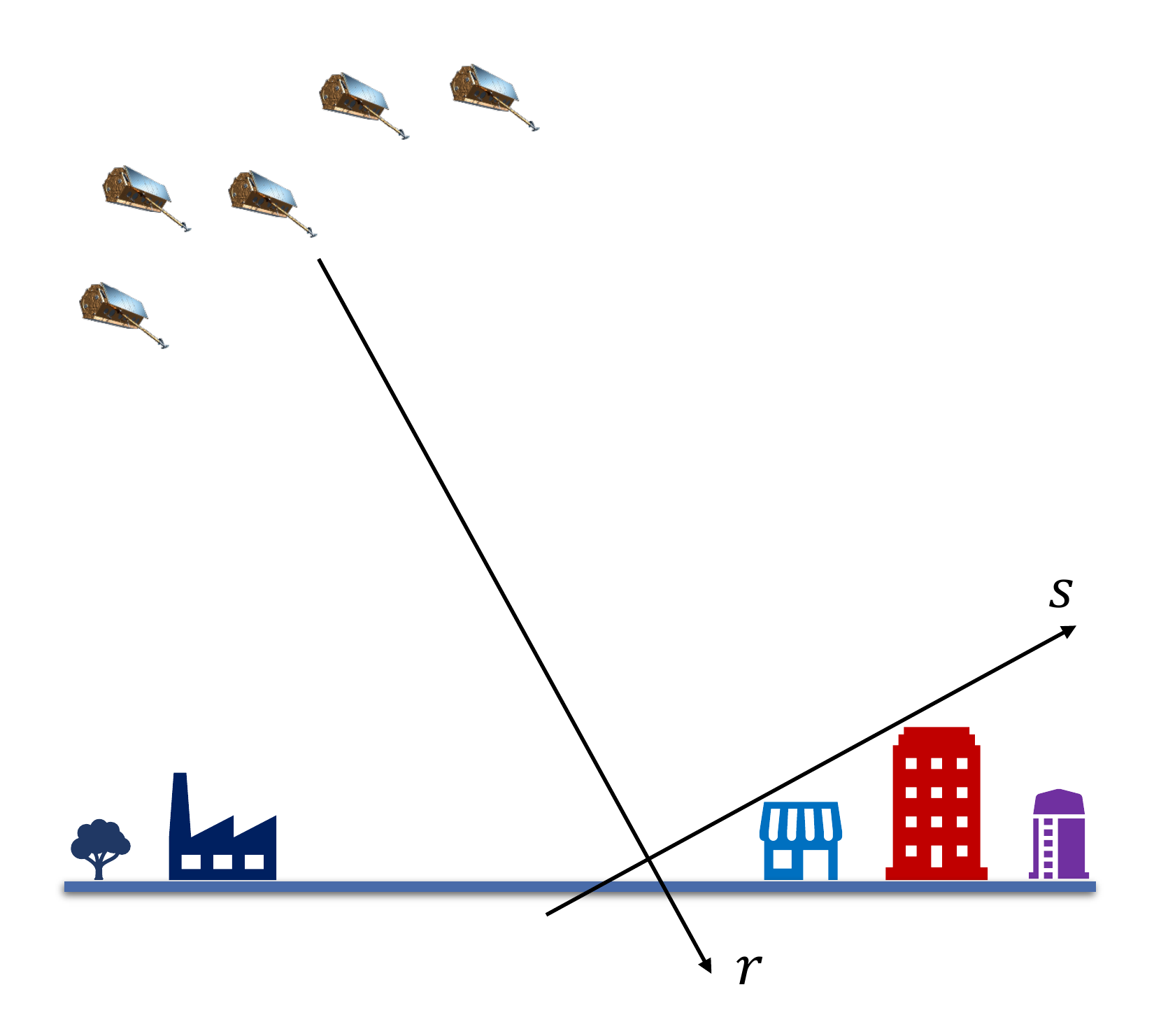}
  \caption{TomoSAR imaging geometry with an artistic view of TerraSAR-X/TanDEM-X}
  \label{fig:nlcs_geometry}
\end{figure}

The typical multi-baseline SAR imaging model can be expressed as follows:
\begin{equation}
g_n = \int_{\Delta s}\gamma(s) \cdot \exp(j2\pi\xi_ns)ds
\end{equation}
where $g_n$ is the complex-valued measurement at an azimuth-range pixel for the $n$th acquisition at time $t_n (n = 1,2,...,N)$. The term $\gamma(s)$ represents the reflectivity function along elevation $s$ with an extent of $\Delta s$. The spatial frequency $\xi_n = 2b_n/\lambda r$ is proportional to the respective aperture position (baseline) $b_n$, $ \lambda $ is the wavelength of the radar signals and $ r $ denotes the range between radar and the observed object, respectively (see Fig. \ref{fig:nlcs_geometry}).

In the presence of noise $\boldsymbol{\varepsilon}$, the discrete-TomoSAR system model can be rewritten as:
\begin{equation}
\mathbf{g}=\mathbf{R}\boldsymbol{\gamma}+\boldsymbol{\varepsilon}
\label{eq:tomosar_basic}
\end{equation}
where $\mathbf{g}$ is the measurement vector with $N$ elements, and $\boldsymbol{\gamma}$ is the reflectivity function along elevation uniformly sampled at $s_l (l=1,2,...,L)$. $\mathbf{R}$ is an $N \times L$ irregularly sampled discrete Fourier transformation mapping matrix. The inherent (Rayleigh) elevation resolution $\rho_s$ of the tomographic arrangement is related to the elevation aperture extent $\Delta b$

\begin{equation}
\rho_s = \dfrac{\lambda r}{2\Delta b}
\end{equation}

\subsection{SL1MMER Algorithm}
To solve (\ref{eq:tomosar_basic}), Zhu et al. proposed ``Scale-down by $L_1$ norm Minimization, Model selection, and Estimation Reconstruction" (SL1MMER) in \cite{bib:Zhu2010a}. They demonstrated its super-resolution power and robustness for spaceborne tomographic SAR in \cite{bib:Zhu2012c}. The SL1MMER algorithm improves the CS algorithm and estimates these parameters in a highly accurate and robust way. It consists of three main steps:
\begin{itemize}
  \item[1)] L1LS Minimization
  \begin{equation}
    \hat{\boldsymbol{\gamma}} = \arg \min_{\boldsymbol{\gamma}} \left\{ \Vert \mathbf{R}\boldsymbol{\gamma} - \mathbf{g} \Vert^2_2 + \lambda \Vert \boldsymbol{\gamma} \Vert_1 \right\}
  \end{equation}
  \item[2)] Model Order Selection
  \begin{equation}
    \hat{K} = \arg \min_{K} \left\{ -2 \ln p \left( \mathbf{g}| \boldsymbol{\theta}\right) + 2 C(K) \right\}
  \end{equation}
  \item[3)] De-aliasing
  \begin{equation}
    \hat{\boldsymbol{\gamma}} = \left( \mathbf{R}^H\mathbf{R} \right)^{-1} \mathbf{R}^H\mathbf{g}
  \end{equation}
\end{itemize}

\begin{figure*}
  \centering
    \includegraphics[width=1.0\textwidth]{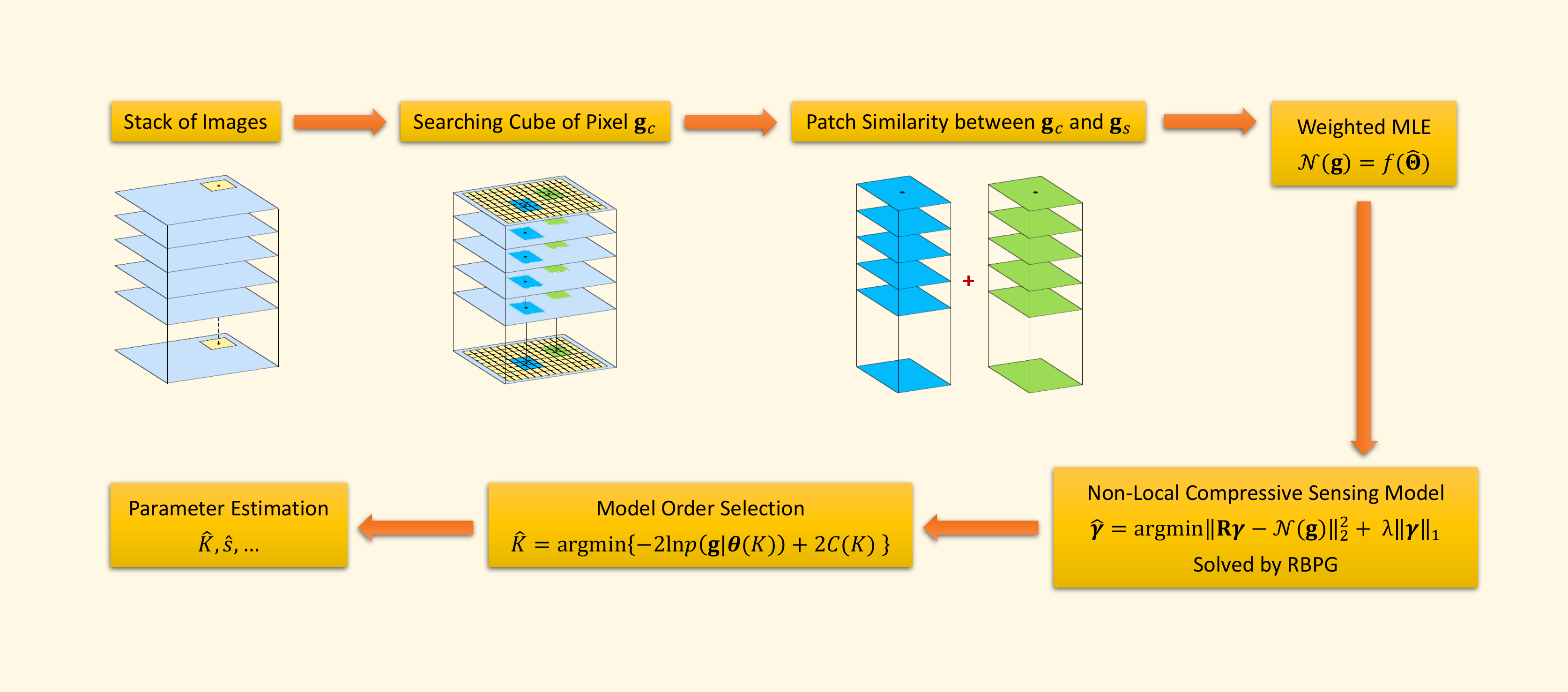}
    \caption{The workflow of Non-Local Compressive Sensing based SAR Tomography}
    \label{fig:nlcs_workflow}
\end{figure*}
where $K$ is the number of scatters. $C(k)$ is a complexity penalty, from which we can see that model selection is actually a tradeoff between how well the model fits the data and
the complexity of the model. $p(\mathbf{g} | \boldsymbol{\theta})$ is the likelihood function which is defined in next section.

Within the framework of SL1MMER, sparse reconstruction and ordinary least squares join forces to incorporate both robust identification of scatterers' elevation positions and accurate amplitude estimation. However, as mentioned above, the CS-based approach has two downsides that can prevent its application. In order to solve those two issues, we need to analyze the estimation accuracy of TomoSAR.

\subsection{Estimation Accuracy}
Assume that $\boldsymbol{\theta}$ is a set of parameters for a given observation $\mathbf{g}$ and $p(\mathbf{g} | \boldsymbol{\theta})$ is the likelihood function. The Cramer-Rao Lower Bound (CRLB) $\mathbf{B}_{CR}$ can be calculated from the inverse of the Fisher information matrix $\mathbf{J}$, which is
\begin{equation}
 \mathbf{B}_{CR} = \mathbf{J}^{-1}
 \end{equation}
 and
\begin{equation}
  \mathbf{J} = -E \left\{ \dfrac{\partial^2 \ln p(\mathbf{g} | \boldsymbol{\theta})}{\partial \boldsymbol{\theta} \partial \boldsymbol{\theta}^H} \right\}
\end{equation}
Since the analytical inversion of $\mathbf{J}$ leads to a complicated expression, the relevant elements of the CRLB matrix can be retrieved by solving the inversion numerically.
The CRLB on elevation estimates of two scatterers can be therefore be defined as
\begin{equation}
  \sigma_{s_q} = c_0 \cdot \sigma_{s_q, 0}
\end{equation}
where
\begin{equation}
  \sigma_{s_q,0} = \dfrac{\lambda r}{4 \pi \cdot \sigma_b \cdot \sqrt{2 \cdot N \cdot \mathrm{SNR} }}
  \label{eq:crlb}
\end{equation}
is the CRLB of the elevation estimates of the $q$th scatterer in the absence of the other one and $\sigma_b$ is the standard deviation of baseline. The essential interference correction factor for closely spaced scatterers is denoted by $c_0$. It has been systematically investigated in \cite{bib:Zhu2012b} that $c_0$ is almost independent of $N$ and SNR, which is defined as:
\begin{equation}
  c_0 = \max \left \{ \sqrt{\dfrac{40\kappa^{-2}(1-\kappa/3)}{9-6(3-2\kappa) \cos(2 \Delta \varphi)+(3-2\kappa)^2}}, 1 \right \}
\end{equation}
where $\kappa = \Delta s / \rho_s$ is the normalized distance between two scatterers and $\Delta \varphi$ is the phase difference.

As shown in Eq. (\ref{eq:crlb}), the estimation accuracy of SL1MMER depends asymptotically on the product $N \cdot \mathrm{SNR}$. Therefore, in order to maintain the estimation accuracy and reduce the number of measurements, SNR needs to be improved. A successful approach to reducing the noise as well as increasing the SNR is the non-local (NL) framework, where the value is a sum weighted with respect to the similarity between the central and other pixels in the search window. NL-means filtering is consistent with the state of the art in image denoising and other applications. Hence, we introduce the NL framework into SL1MMER to achieve good performance with a minimal number of acquisitions.

\section{Non-Local CS-Based SAR Tomography}
The non-local concept proposed in \cite{bib:buades2005non} takes advantage of the high degree of redundancy of any natural image. This means that every feature edge, point, etc. in an image can be found similarly many times in the same image. Inspired by the neighborhood filters, such as boxcar and adaptive filters, the NL-means concept redefines the neighborhood of a pixel $c$ in a very general sense as any set of pixels $s$ in the image (local or non-local) such that a small patch around $s$ is similar to the patch around $c$. Fig. \ref{fig:nlcs_workflow} shows the workflow of the proposed non-local compressive sensing based tomographic SAR (NLCS-TomoSAR) method.

\subsection{Non-Local Compressive Sensing}
In cases where there is no prior knowledge about the number of scatters and in the presence of measurement noise, the non-local CS-based TomoSAR inversion can be written as
\begin{equation}
\hat{\boldsymbol{\gamma}} = \arg \min_{\boldsymbol{\gamma}} \{ \Vert \mathbf{R}\boldsymbol{\gamma} - \mathcal{N}(\mathbf{g}) \Vert^2_2 + \lambda \Vert \boldsymbol{\gamma} \Vert_1 \}
\label{equ:opt_nll1lsp}
\end{equation}
where $\mathcal{N}(.)$ is the non-local estimator and $\mathcal{N}(\mathbf{g}) = f( \hat{\boldsymbol{\Theta}})$.  The expression $\hat{\boldsymbol{\Theta}} = (\hat{\psi}, \hat{\mu}, \hat{\sigma^2})$ denotes the parameters, where $\hat{\psi}$ is the estimate of the interferometric phase, $\hat{\mu}$ is the coherence magnitude, and $\hat{\sigma^2}$ is variance, which will be introduced later. NL-means can combine similar patches into a weighted maximum likelihood estimator (WMLE)
\begin{equation}
\hat{\boldsymbol{\Theta}}_c = \mathrm{argmax} \sum_s \mathbf{w}(i_s, j_s) \log p(\mathbf{g}_s|\boldsymbol{\Theta})
\end{equation}
The measure of the patch similarity that leads to the weights $\mathbf{w}(i_s, j_s)$ depends on the statistical model of the imaging process. In our case it is derived from the InSAR statistics.

\subsection{Interferometric SAR statistics}
The underlying statistical model for a fully developed speckle field is that of a circular complex Gaussian random process that yields the M-dimensional Gaussian probability density function (PDF) \cite{bib:goodman2007speckle}.
\begin{equation}
  p(\mathbf{g}|\mathbf{C}) = \dfrac{1}{\pi^{M}|\mathbf{C}|} \exp(-\mathbf{g}^{\mathrm{H}} \mathbf{C}^{-1} \mathbf{g})
\end{equation}
where $\mathbf{C}$ is the covariance matrix. A special case of interest is $M=2$ for InSAR, which leads to the simplified form for the joint PDF of $g(I_1, I_2, \phi)$:
\begin{multline}
  p(I_1, I_2, \phi) = \dfrac{1}{16\pi^2 \sigma^4(1-\mu^2)}  \\ \times \exp \left[-\dfrac{I_1 + I_2 - 2 \sqrt{I_1 I_2}\mu \cos(\phi - \psi)}{2\sigma^2(1-\mu^2)} \right]
  \label{eq:pdf_insar}
\end{multline}
where $I_1$ and $I_2$ are intensities of two coregistered SAR images, and it has been assumed that $\langle I_1 \rangle = \langle I_2 \rangle = 2 \sigma^2$. $\phi$ is the noisy interferometric phase. By imposing a scale-invariant similarity criterion, the weight is set as a function of likelihood:
\begin{equation}
  \mathbf{w}(i_s, j_s) = \prod_m p(I_{1,m}, I_{2,m}, \phi_m)^{1/h}
\end{equation}
where $h$ is a filtering parameter, the same as in \cite{bib:deledalle2011nl}. By applying the maximum likelihood estimation of Eq. (\ref{eq:pdf_insar}) derived in \cite{bib:seymour1994maximum}, the estimated parameters can be formulated as
\begin{eqnarray}
  \hat{\psi}_{wmle}  &=& -\arg \left(\sum_s \mathbf{w}_s \mathbf{g}_{1,s} \mathbf{g}_{2,s}^{*} \right) \\
  \hat{\mu}_{wmle}   &=& \dfrac{2 \sum_s \mathbf{w}_s |\mathbf{g}_{1,s}| |\mathbf{g}_{2,s}|}{\sum_s \mathbf{w}_s \left(|\mathbf{g}_{1,s}|^2 + |\mathbf{g}_{2,s}|^2 \right)} \\
  \hat{\sigma^2}_{wmle} &=& \dfrac{\sum_s \mathbf{w}_s \left(|\mathbf{g}_{1,s}|^2 + |\mathbf{g}_{2,s}|^2 \right)}{4 \sum_s \mathbf{w}_s}
\end{eqnarray}

\section{An Efficient Algorithm for solving the NLCS model}
Non-local filtering and sparse reconstruction algorithms are usually computationally expensive and are difficult to extend to large scales. In this section, we introduce an approach for solving the NLCS model, which can retain the super-resolution power of the standard basis pursuit denoising (BPDN) solver and considerably speed up the processing for matrix \textbf{R} of the random Fourier transform, as used in SL1MMER.

\subsection{Optimized Parallelization of Non-Local Process}
Note that pixels outside searching windows do not contribute to the value of the central pixel in a non-local process. This property allows us to separate the image into independent disjoint pieces and process them in parallel, as it is done in domain decomposition schemes. In \cite{bib:shi2015optimized}, we proposed a sophisticated and optimized parallelization scheme for non-local processing. A message passing interface (MPI) was adopted for non-local processes, enabling us to use thousands of cores for large-scale processing. The bottleneck of this process is the communication between cores. We introduced a synchronized communication scheme to avoid the bottleneck and the speedup increased dramatically with the increase in the number of cores.

\subsection{Fast and Accurate Solver for $L_1$ least squares minimization}
In SL1MMER, the second order method primal-dual interior-point method (PDIPM) with self-dual embedding technique was adopted to solve the second order cone program, which is extremely expensive with respect to computation. The algorithm proposed in \cite{bib:shi2018fast}, ``randomized blockwise proximal gradient (RBPG)'', splits the unconstrained optimization problems into two parts, the convex differentiable part and the convex non-differentiable part, leading to the so-called proximal gradient (PG) method. The iterative approach to solve Eq. (\ref{equ:opt_nll1lsp}) can be written as
\begin{equation}
{{\boldsymbol{\gamma}}^{k+1}}= \arg \min \Big( \langle \nabla f({\boldsymbol{\gamma}}^k), {\boldsymbol{\gamma}}-{\boldsymbol{\gamma}}^k \rangle + \frac{1}{2\alpha_k}\Vert {\boldsymbol{\gamma}}-{\boldsymbol{\gamma}}^k \Vert_2^2 + r(\boldsymbol{\gamma}) \Big)
\label{equ:opt_sol}
\end{equation}
where $f$ is $\Vert \mathbf{R}\boldsymbol{\gamma} - \mathcal{N}(\mathbf{g}) \Vert^2_2$ and $\nabla f$ is the partial gradient of function $f$. The proximal gradient formulation is
\begin{equation}
{\boldsymbol{\gamma}}^{k+1} ={\rm prox}_{\alpha_kr}({\boldsymbol{\gamma}}^k - \alpha_k\nabla f({\boldsymbol{\gamma}}^k))
\end{equation}
where $\alpha_k > 0$ is step size, can be constant or determined by line search. For $r({\boldsymbol{\gamma}}) = \Vert {\boldsymbol{\gamma}} \Vert_1$, the proximal operator can be chosen as soft-thresholding.

After applying Nesterov's acceleration scheme and block coordinate techniques, the equation (\ref{equ:opt_sol}) can be written as
\begin{eqnarray}
{\boldsymbol{\gamma}}_{i_k}^{k+1} = &\arg \min \Big(\langle \nabla f_{i_k}({\boldsymbol{\gamma}}_{i_k}^k), {\boldsymbol{\gamma}}_{i_k}-{\boldsymbol{\gamma}}_{i_k}^k \rangle \nonumber \\
&+ \frac{1}{2\alpha_{i_k}^k}\Vert {\boldsymbol{\gamma}}_{i_k}-{\boldsymbol{\gamma}}_{i_k}^k \Vert_2^2 + r_{i_k}(\boldsymbol{\gamma}) \Big)
\end{eqnarray}
where $i_k$ is the index of a block. The choice of the update index $i_k$ for each iteration is crucial for good performance. Often, it is easy to switch index orders. However, the choice of index affects convergence, possibly resulting in faster convergence or divergence. In this work, we choose a randomized variants scheme, whose strengths include less memory consumption, good convergence performance, and empirical avoidance of the local optima. The block index  $i_k$ is chosen randomly following the probability distribution given by the vector
\begin{equation}
P_{i_k} = \dfrac{L_{i_k}}{\sum_{j=1}^J L_j}, \quad i_k = 1,...,J
\end{equation}
where $L_{i_k}$ is the Lipschitz constant of $\nabla_{i_k}f({\bf x})$, the gradient of $f({\bf x})$ with respect to the $i_k$-th group (in our case $L = ||\mathbf{R}^T\mathbf{R}||$). However, setting $\alpha_k = 1/L$ usually results in very small step sizes; hence, the time step $\alpha_k$ is adaptively chosen by using the backtracking line search method.

\begin{figure*}
\centering
\subfloat[]{\includegraphics[width=0.3\textwidth]{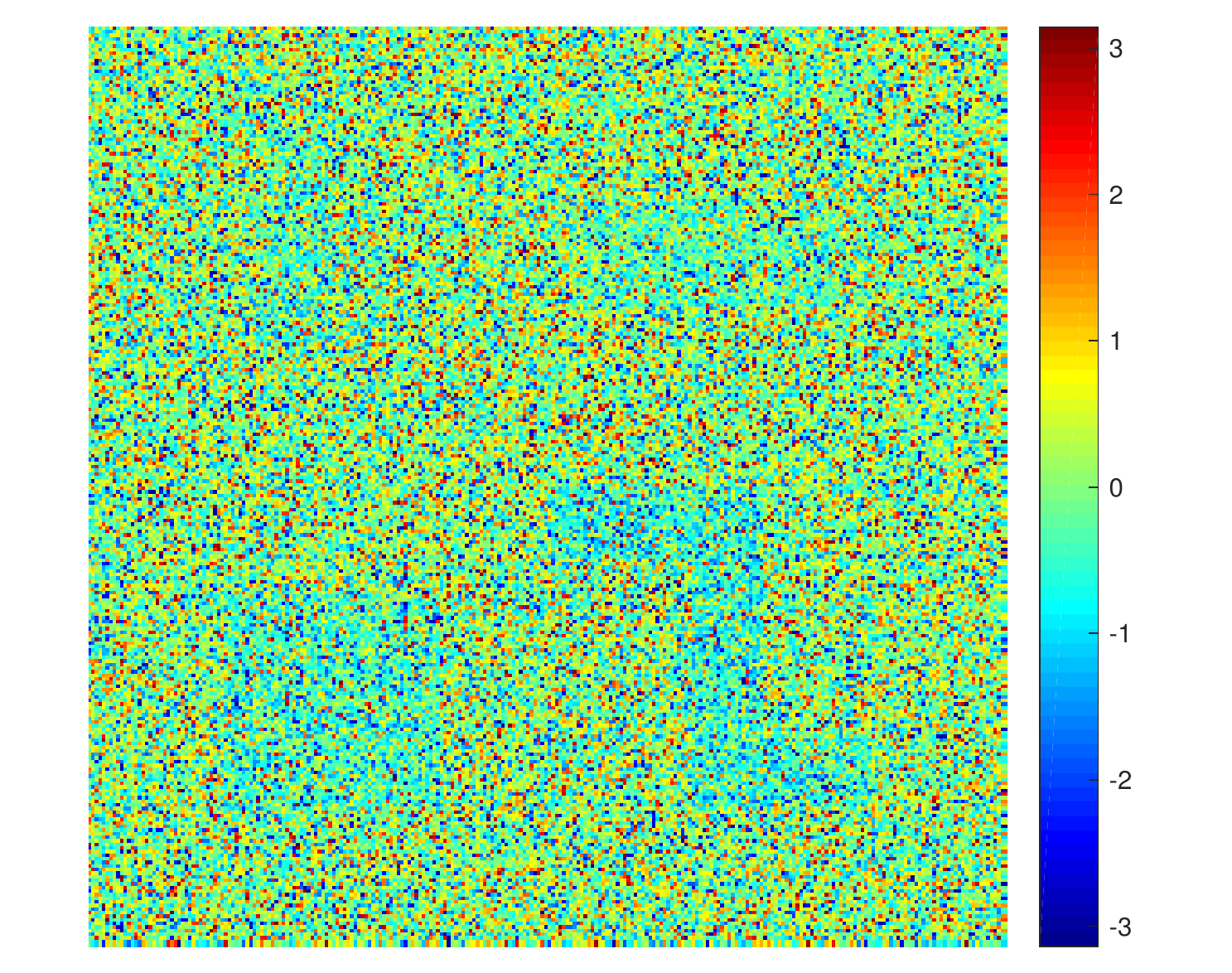}}
\subfloat[]{\includegraphics[width=0.3\textwidth]{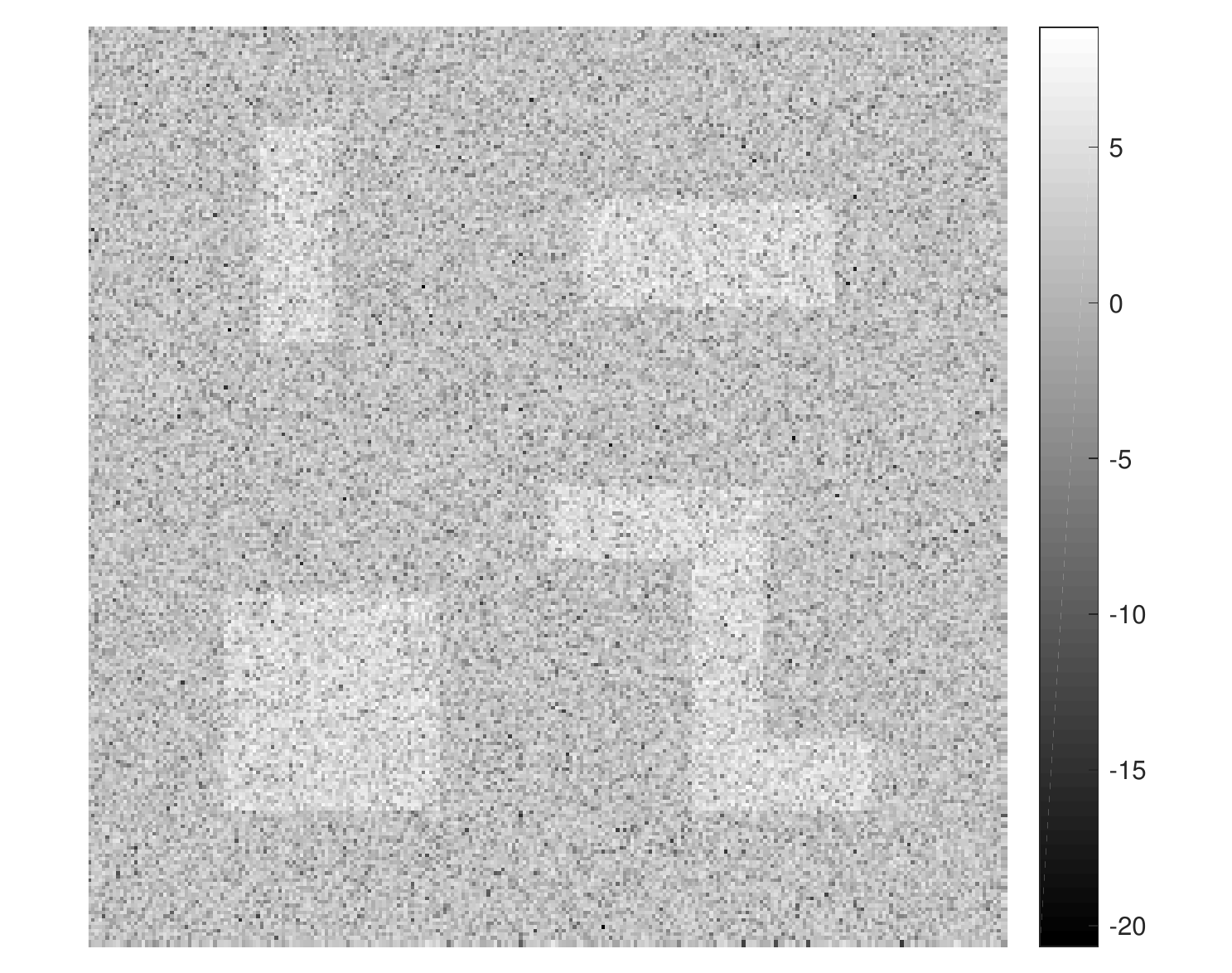}}
\subfloat[]{\includegraphics[width=0.3\textwidth]{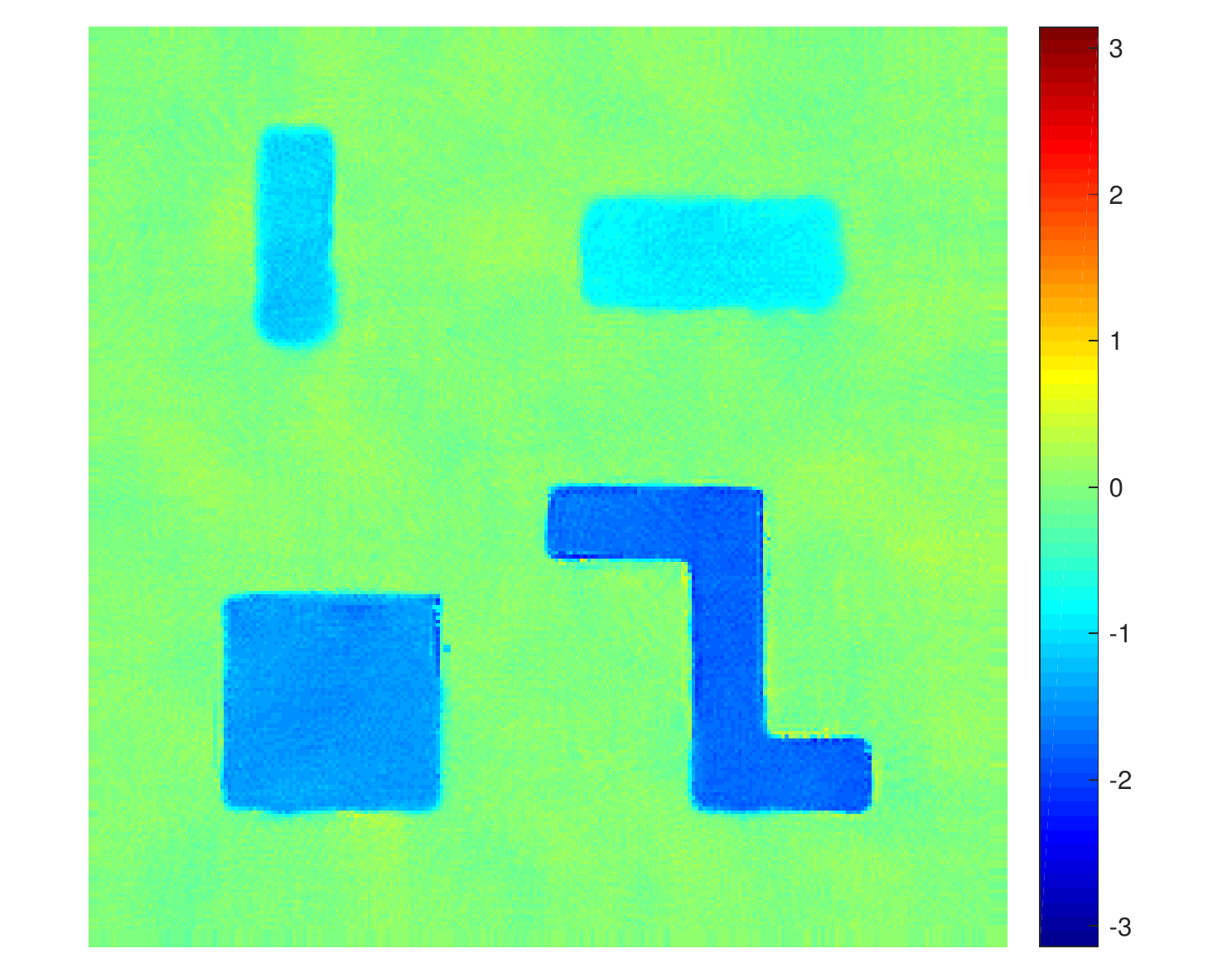}}
\hfil
\subfloat[]{\includegraphics[width=0.3\textwidth]{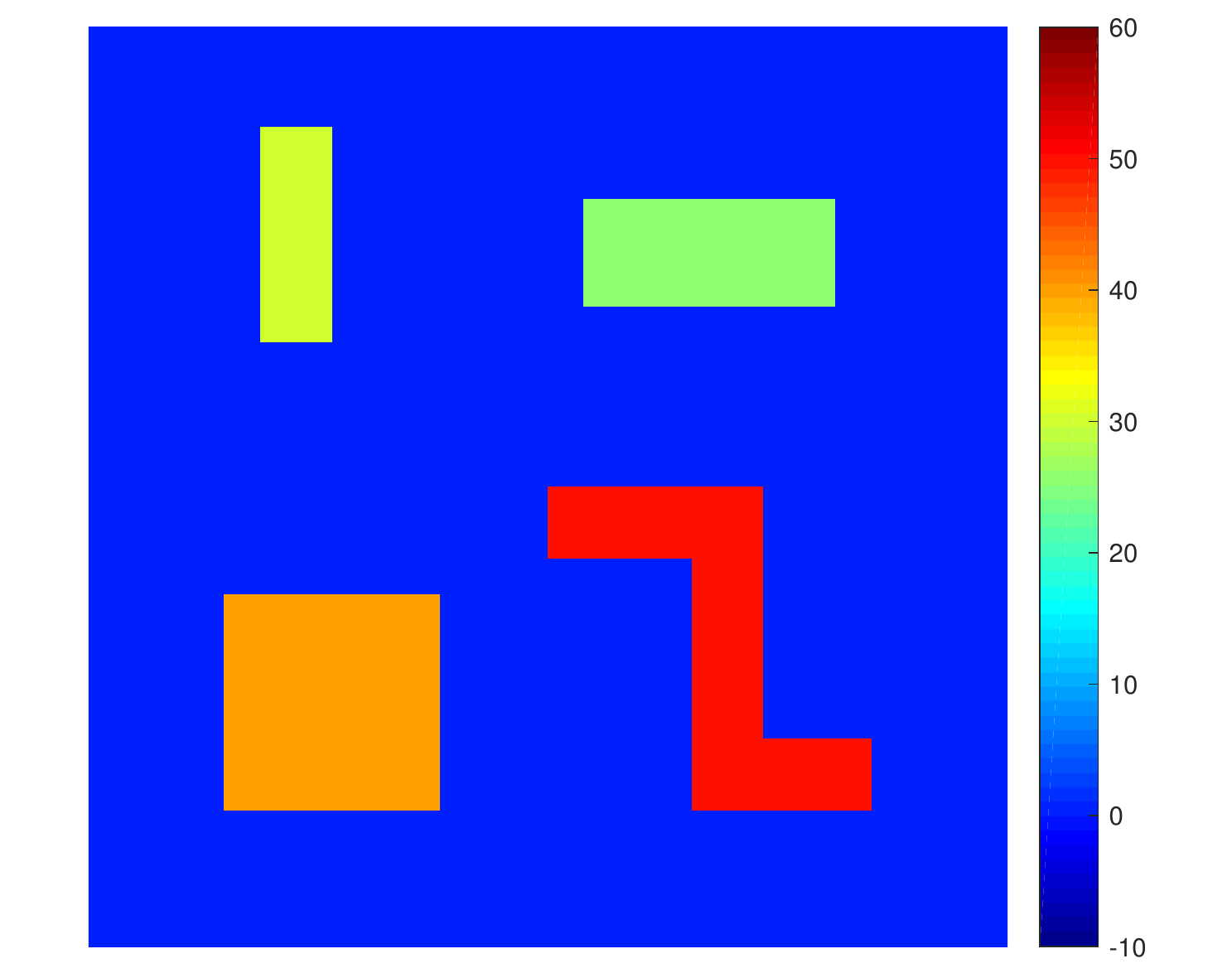}}
\subfloat[]{\includegraphics[width=0.3\textwidth]{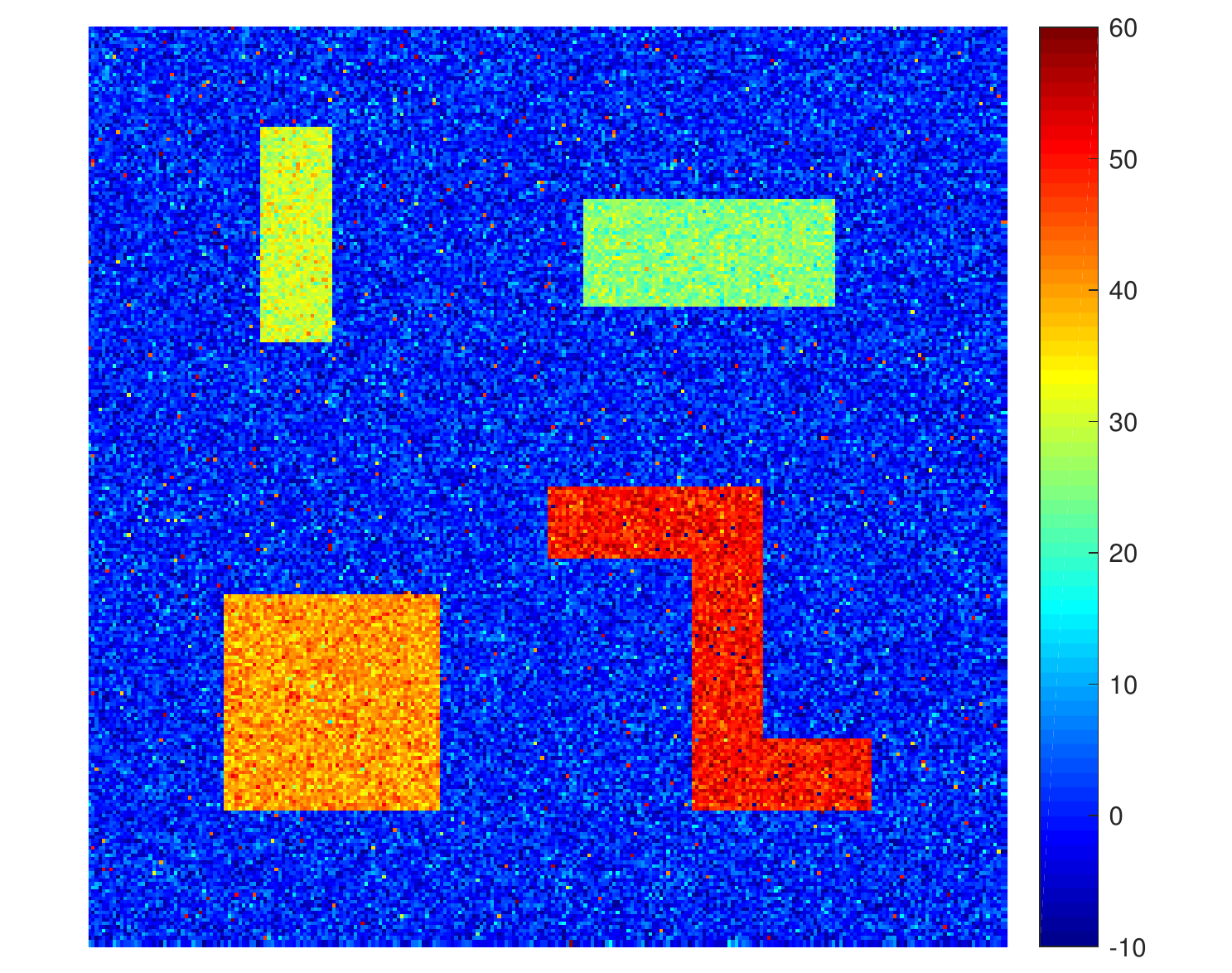}}
\subfloat[]{\includegraphics[width=0.3\textwidth]{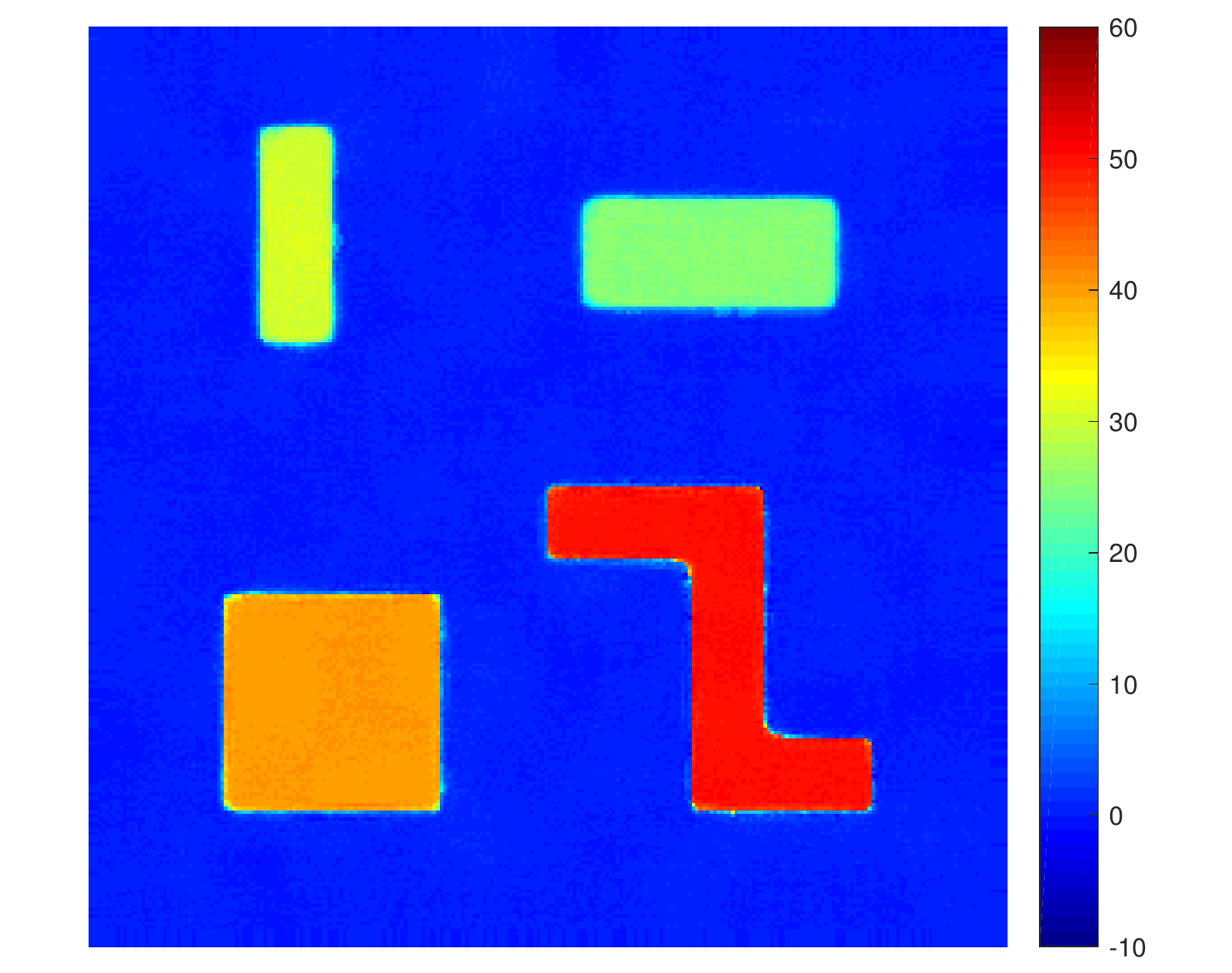}}
\caption{Simulated data with SNR = 3 dB. (a) One interferogram. (b) Corresponding amplitude. (c) Filtered interferogram. (d) Ground truth of height. (e) Reconstructed height by TomoSAR. (f) Reconstructed height by NLCS-TomoSAR.}
\label{fig:experiments_simData_p0p2}
\end{figure*}

\section{Numerical Results with Simulated Data}
In this section, we perform our proposed NLCS approach on simulated data. The simulated complex data have been generated from the height profile and different SNR. An urban-like scene has been generated with rectangular geometric shapes.

The characteristics of the profile and of the scene are reported in Table \ref{tab:simData_char}.
\begin{table}[h!]
\begin{center}
\caption{Parameters of Numerical Simulation}
\label{tab:simData_char}
\begin{tabular}{lll}
\toprule
Shape & Height (meter) & Size (pixels)\\
\midrule
Top left & $30$ m & $60 \times 20$\\
Top right & $25$ m & $30 \times 70$\\
Bottom left & $40$ m & $60 \times 60$\\
Bottom right & $50$ m & $20 \times 50 \times 2 + 20 \times 60$\\
\bottomrule
\end{tabular}
\end{center}
\end{table}
The noise-free phase of each interferogram was calculated by using a realistic TerraSAR-X baseline distribuation with 29 interferograms.
\begin{equation}
  \phi_{sim} = \dfrac{4\pi b_n h}{\lambda r \sin \theta_{inc}}
  \label{eq:h2p}
\end{equation}
Two stacks of complex data were generated with SNR = 3, -8 dB.

Fig. \ref{fig:experiments_simData_p0p2} shows the example of one interferogram (a) and its corresponding amplitude (b), the interferogram after non-local filtering (c), the ground truth of the buildings' height (d), the reconstructed height by TomoSAR (e), and the reconstructed height by NLCS-TomoSAR (f). As is apparent, the estimation of height by the original TomoSAR can give an acceptable result when the SNR is relatively high. Compared to the original TomoSAR, NLCS-TomoSAR shows a more accurate result and small loss of resolution at the edges.

\begin{figure*}
\centering
\subfloat[]{\includegraphics[width=0.3\textwidth]{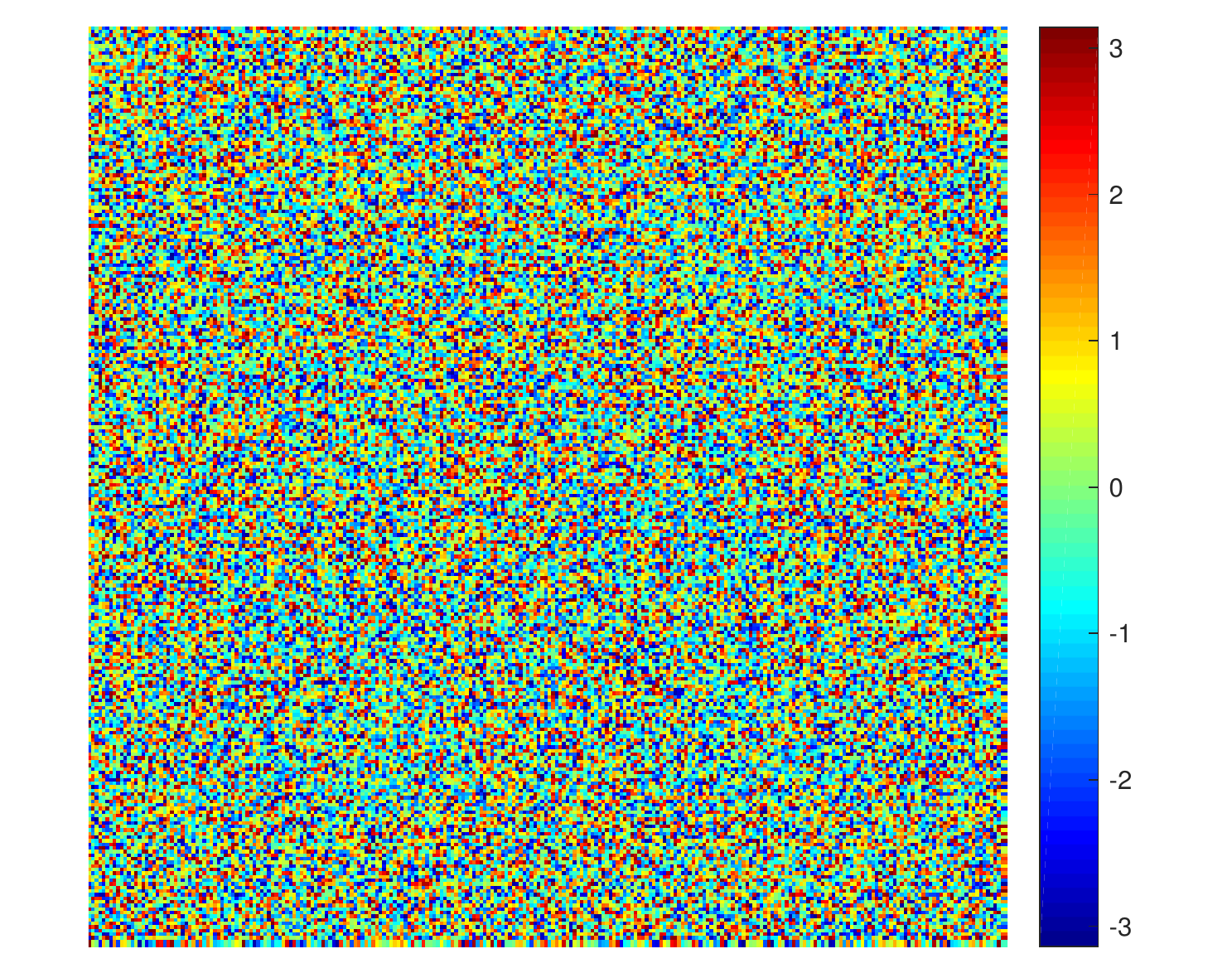}}
\subfloat[]{\includegraphics[width=0.3\textwidth]{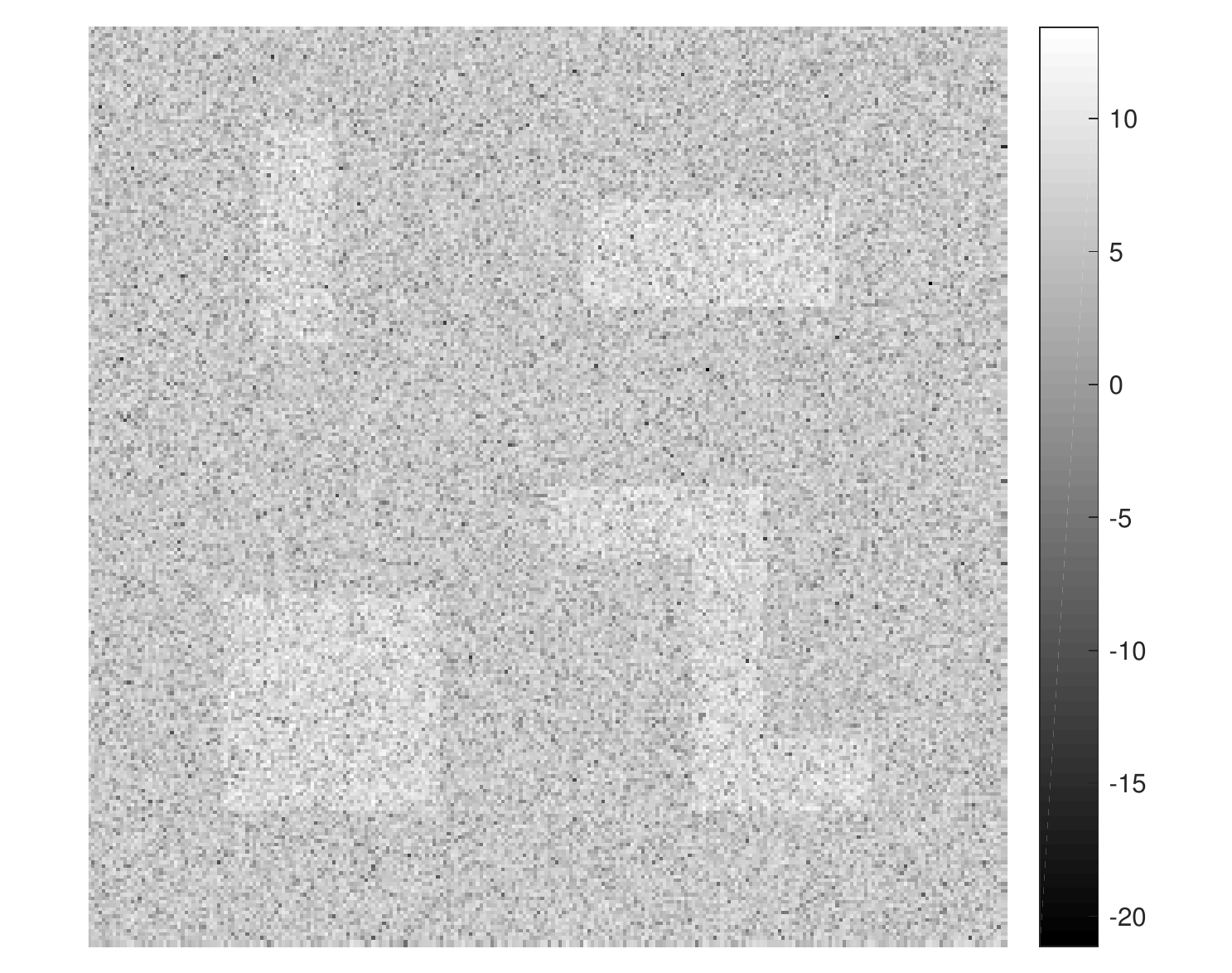}}
\subfloat[]{\includegraphics[width=0.3\textwidth]{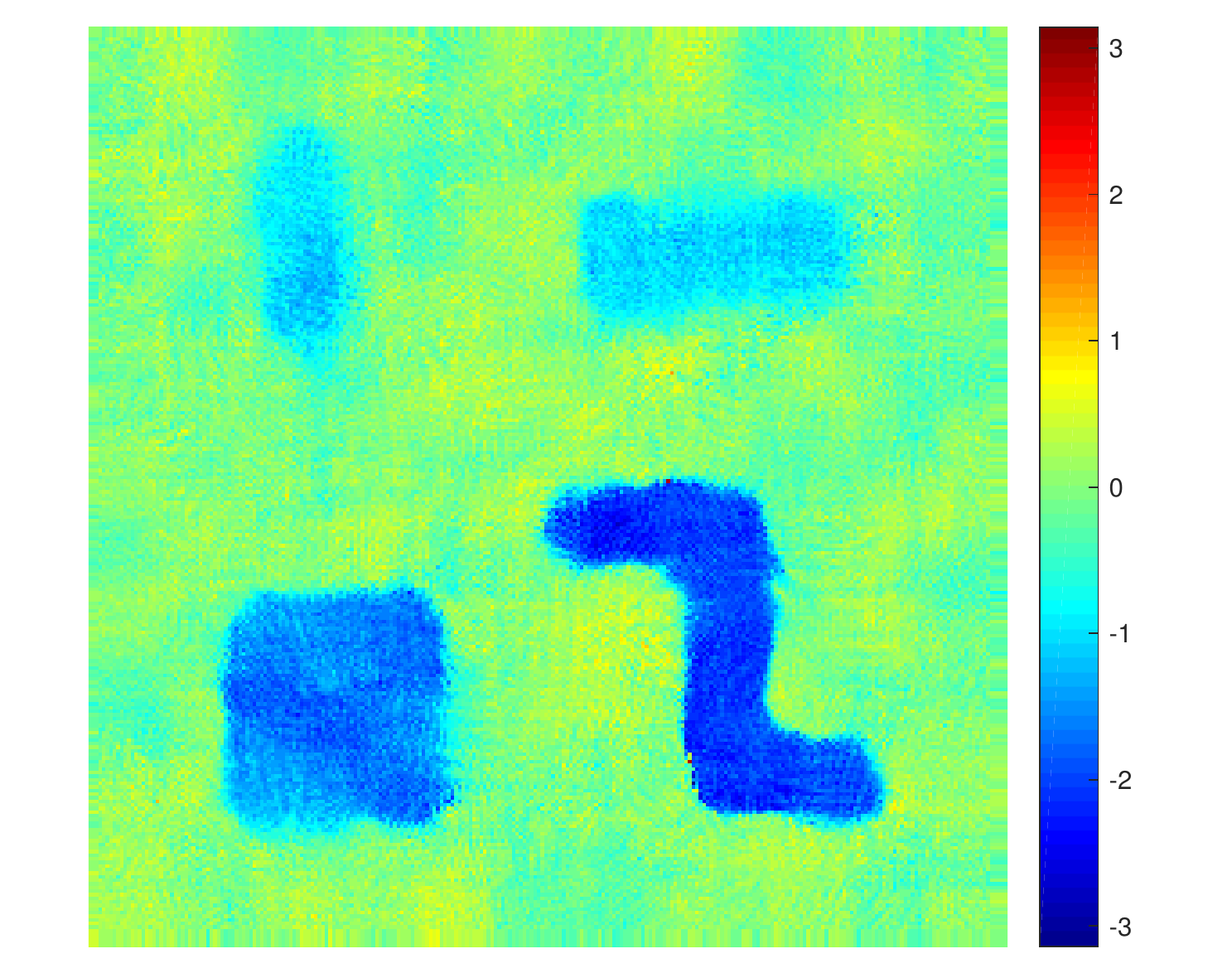}}
\hfil
\subfloat[]{\includegraphics[width=0.3\textwidth]{nltomosar_gt_sim}}
\subfloat[]{\includegraphics[width=0.3\textwidth]{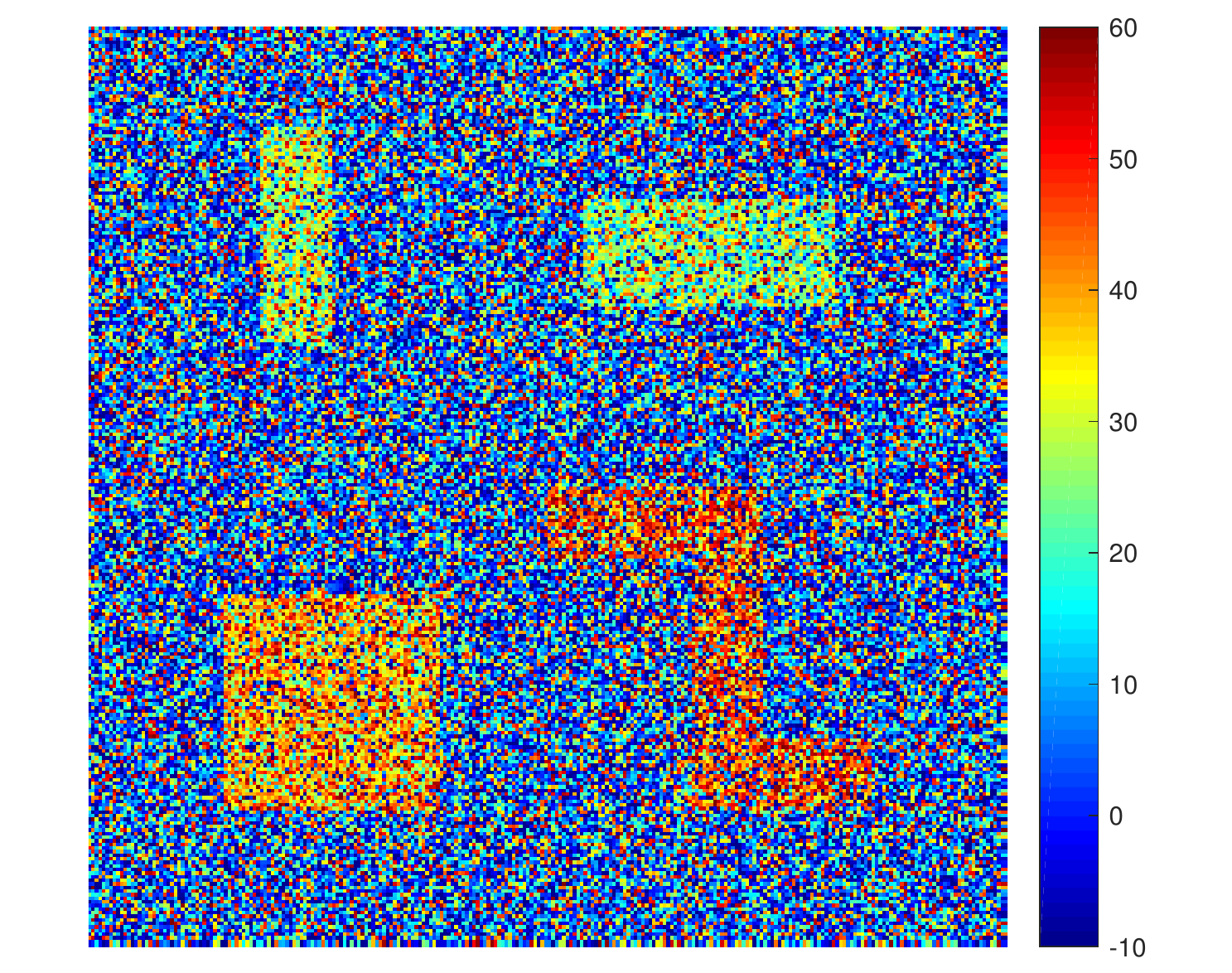}}
\subfloat[]{\includegraphics[width=0.3\textwidth]{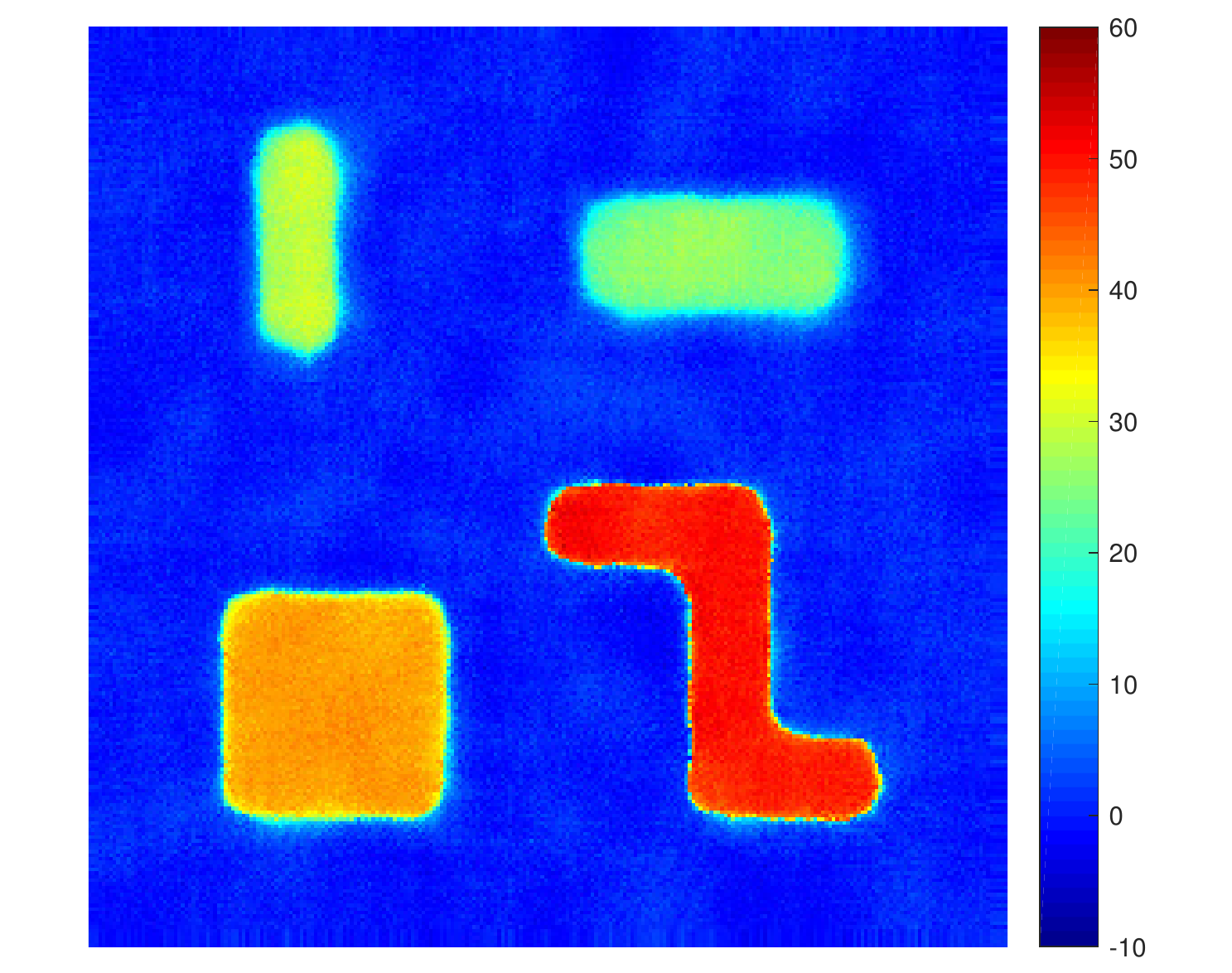}}
\caption{Simulated data with SNR = -8 dB. (a) One interferogram. (b) Corresponding amplitude. (c) Filtered interferogram. (d) Ground truth of height. (e) Reconstructed height by TomoSAR. (f) Reconstructed height by NLCS-TomoSAR.}
\label{fig:experiments_simData_n10n8}
\end{figure*}
Fig. \ref{fig:experiments_simData_n10n8} shows the result of the same configuration as Fig. \ref{fig:experiments_simData_p0p2}, but with a different SNR = -8 dB. It can be seen that the interferogram in Fig. \ref{fig:experiments_simData_n10n8} (a) is strongly blurred and the pattern cannot be easily recognized. After applying the NL filter, the structure of the buildings is visible in Fig. \ref{fig:experiments_simData_n10n8} (c). The height estimation produced very noisy estimates by TomoSAR for low SNR and the accuracy of the estimates is quite low. In contrast, the estimates of the height by NLCS are extremely good. There is only resolution loss at the edges due to very low SNR.

\begin{figure*}
\centering
\subfloat[]{\includegraphics[width=0.3\textwidth]{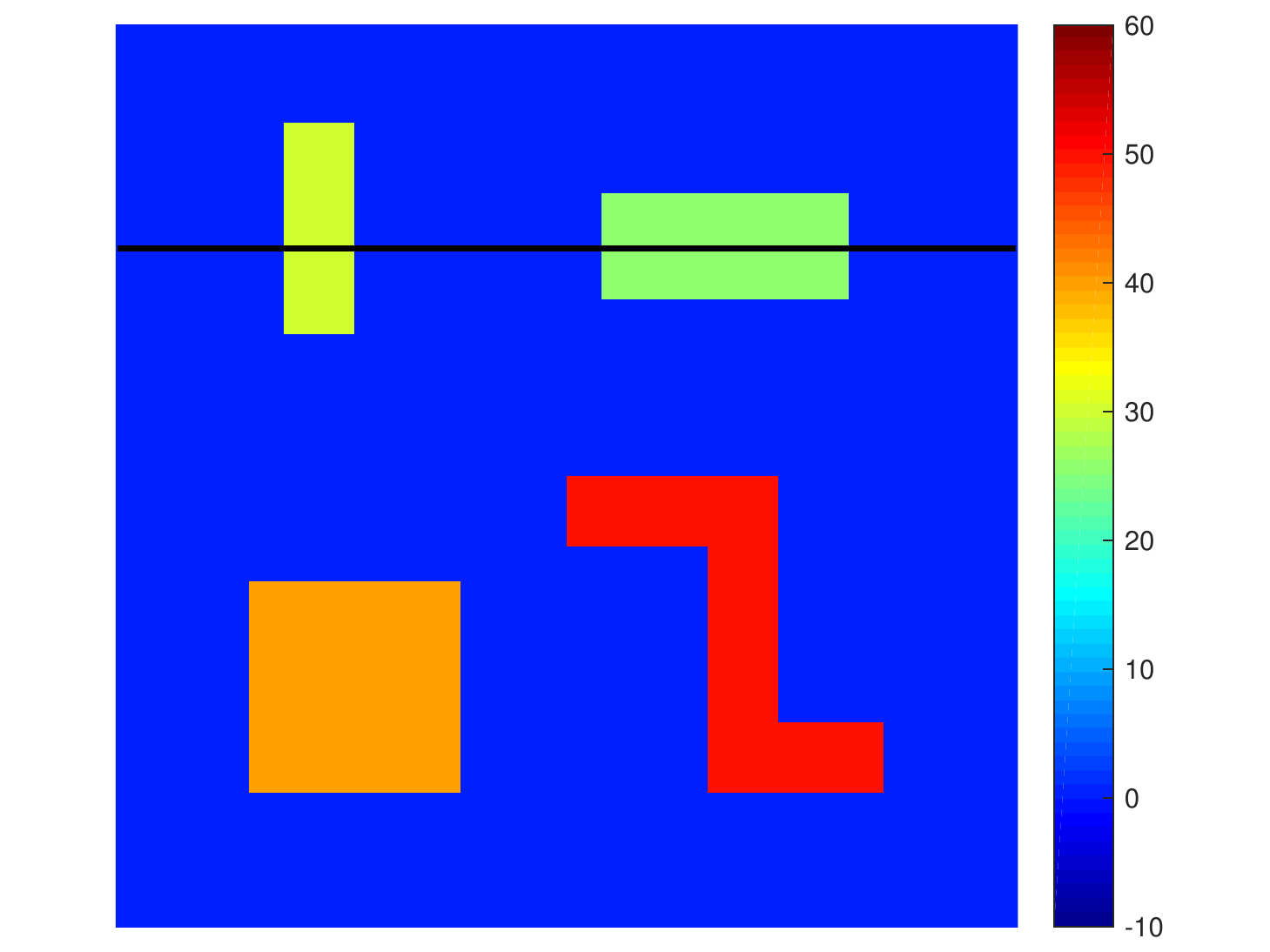}}
\subfloat[]{\includegraphics[width=0.3\textwidth]{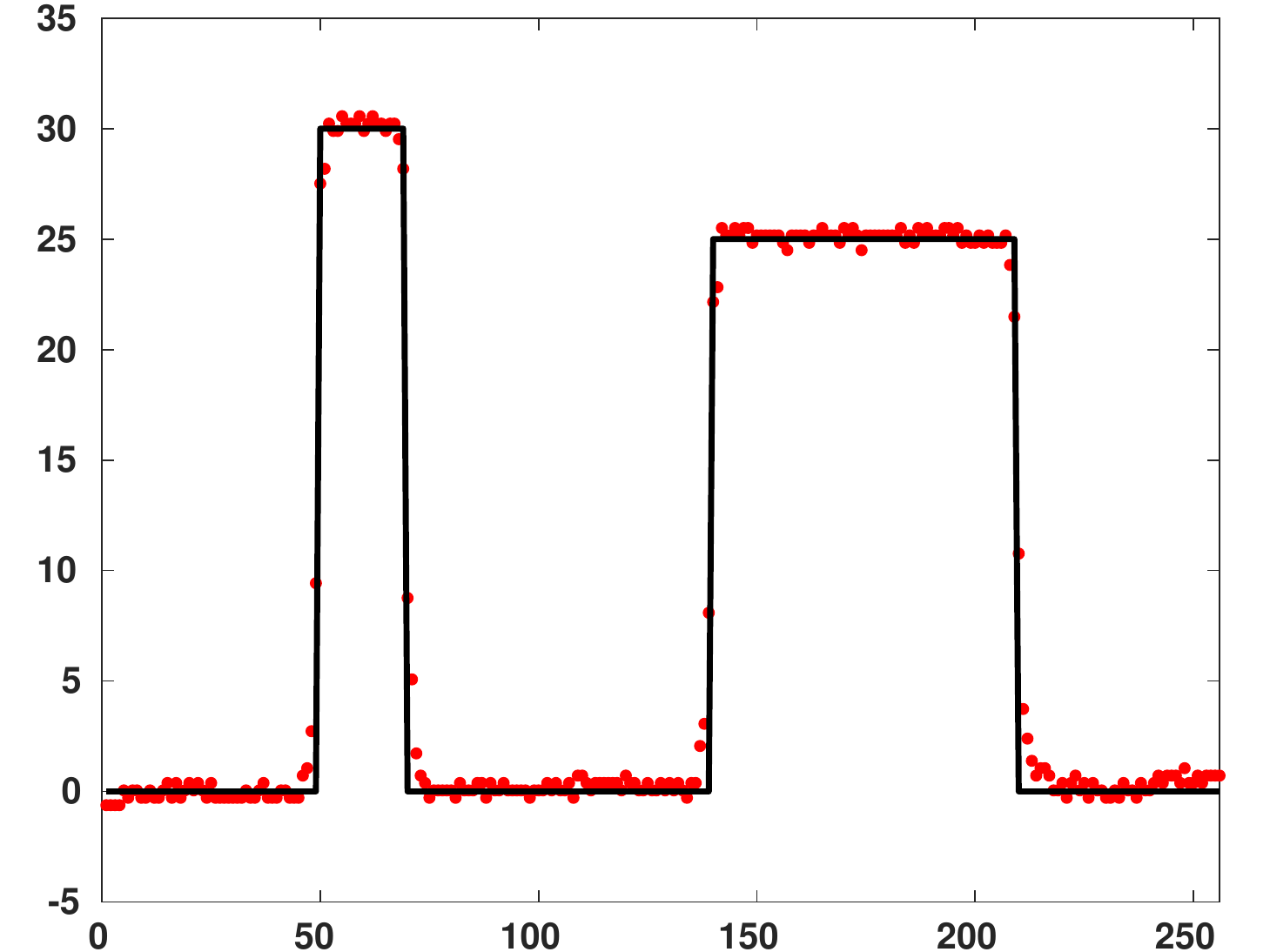}}
\subfloat[]{\includegraphics[width=0.3\textwidth]{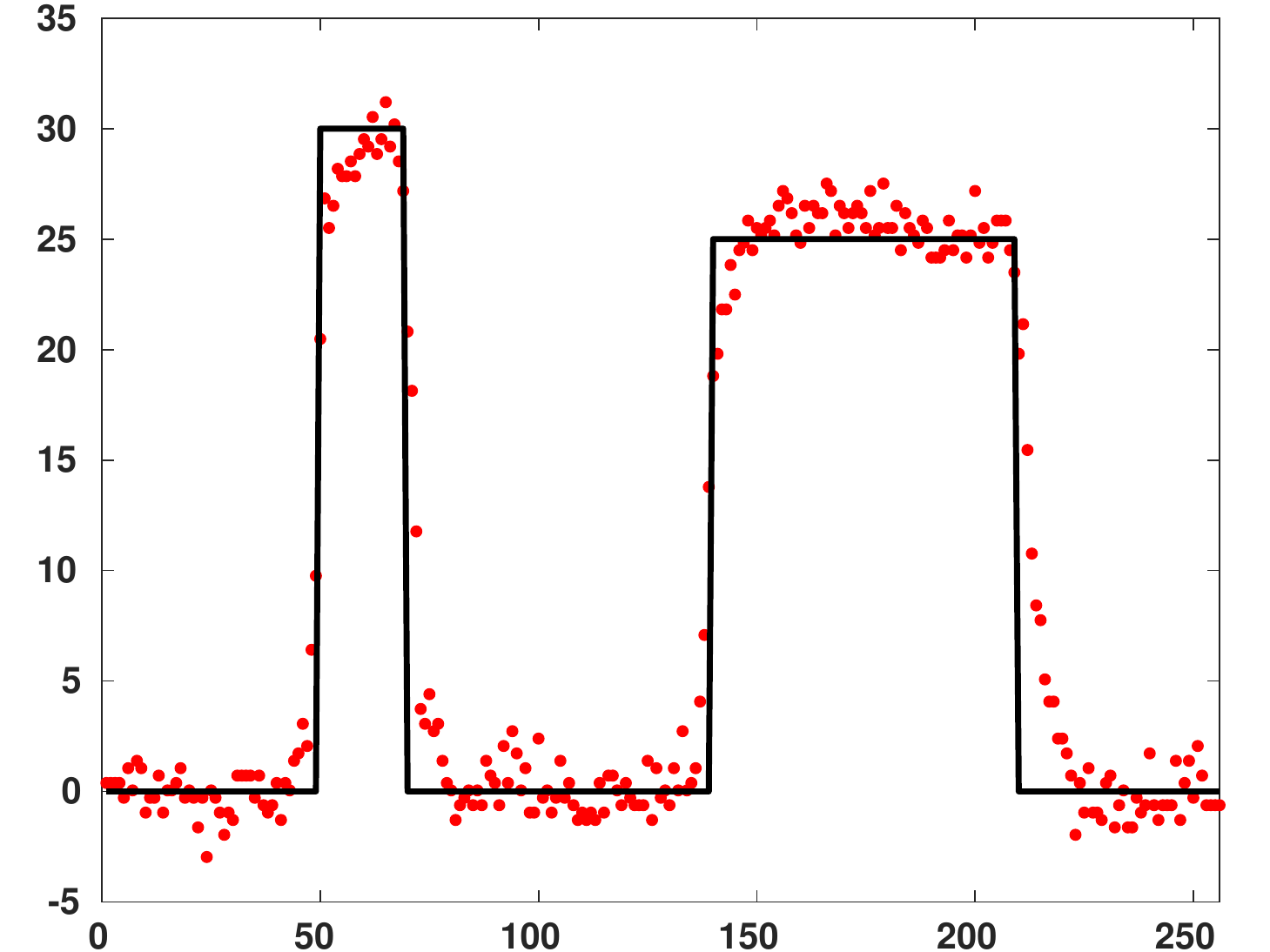}}
\caption{Height profile of reconstructed result. (a) Position of height profile. (b) Height profile of the case (SNR = 3 dB). (c) Height profile of the case (SNR = -8 dB). Solid black line denotes the ground truth of the height and red dots represents the estimated height by proposed method.}
\label{fig:experiments_simData_slice}
\end{figure*}

Fig. \ref{fig:experiments_simData_slice} presents the slice of height profile along with the solid black line for both SNR = 3 and -8 dB. The black line in Fig. \ref{fig:experiments_simData_slice} (b) and (c) is the ground truth of the height profile and the red dots are the height in each pixel estimated by NLCS-TomoSAR. It can be seen that, for relatively high SNR (3 dB), the spatial bias is not notable and variance is quite small. The standard deviation is 0.23 m, 0.24 m, and the mean error is 0.21 m, 0.15 m for shape 1 and shape 2, respectively. For the low SNR case, both the spatial bias and variance increase. The detailed results are shown in Table \ref{tab:simData_height}.

\begin{table}
\begin{center}
\caption{Statistics of Height Estimation}
\label{tab:simData_height}
\begin{tabular}{lccc}
\toprule
 & Ground Truth & Mean Value & Standard Deviation\\
\midrule
Shape 1 (3dB) & 30 m & 30.21 m & 0.23 m\\
Shape 2 (3dB) & 25 m & 25.15 m & 0.24 m\\
Shape 1 (-8dB) & 30 m & 29.21 m & 1.44 m\\
Shape 2 (-8dB) & 25 m & 25.43 m & 1.16 m\\
\bottomrule
\end{tabular}
\end{center}
\end{table}

\section{Practical Demonstration with TerraSAR-X Data}
In this section, we evaluate the performance of the proposed method using real TerraSAR-X data.
\subsection{Data Description}
The test area in this work is the headquarters of the German Railway, the ``Deutsche Bahn'' (DB) in Munich. We chose TerraSAR-X high resolution spotlight data with a slant-range resolution of 0.6 m and an azimuth resolution of 1.1 m, which consists of 64 interferograms in one stack acquired with a range bandwidth of 300 MHz. 
The elevation aperture size $\Delta b$ is about 254.07 m. The detailed parameters of TerraSAR-X acquisition are shown in Table \ref{tab:realData_char}. The preprocessing including atmosphere phase screen correction was performed by the German Aerospace Center (DLR) PSI-GENESIS system on a persistent scatterer network of high-SNR pixels containing only single scatterers \cite{bib:adam2008high}. The SL1MMER algorithm with Bayesian information criterion \cite{bib:schwarz1987estimating} as the model selection scheme was applied to each pixel of the test area.

\begin{table}
\begin{center}
\caption{TerraSAR-X Acquisition Parameters}
\label{tab:realData_char}
\begin{tabular}{cccc}
\toprule
$r$ & $\lambda$ & $\theta_{inc}$ & $\Delta b$\\
\midrule
704 km & $3.1$ cm & $39.36^{\circ}$ & 254.07 m\\
\bottomrule
\end{tabular}
\end{center}
\end{table}

\subsection{Experimental Results}
In order to compare the performance of different algorithms, we extracted two new stacks from the original 64 images with 7 images and 14 images, respectively.

\begin{figure*}
  \centering
  \subfloat[]{\includegraphics[width=0.48\textwidth]{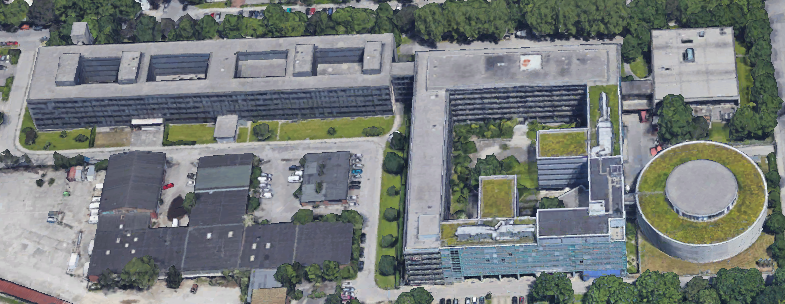}}
  \subfloat[]{\includegraphics[width=0.5\textwidth]{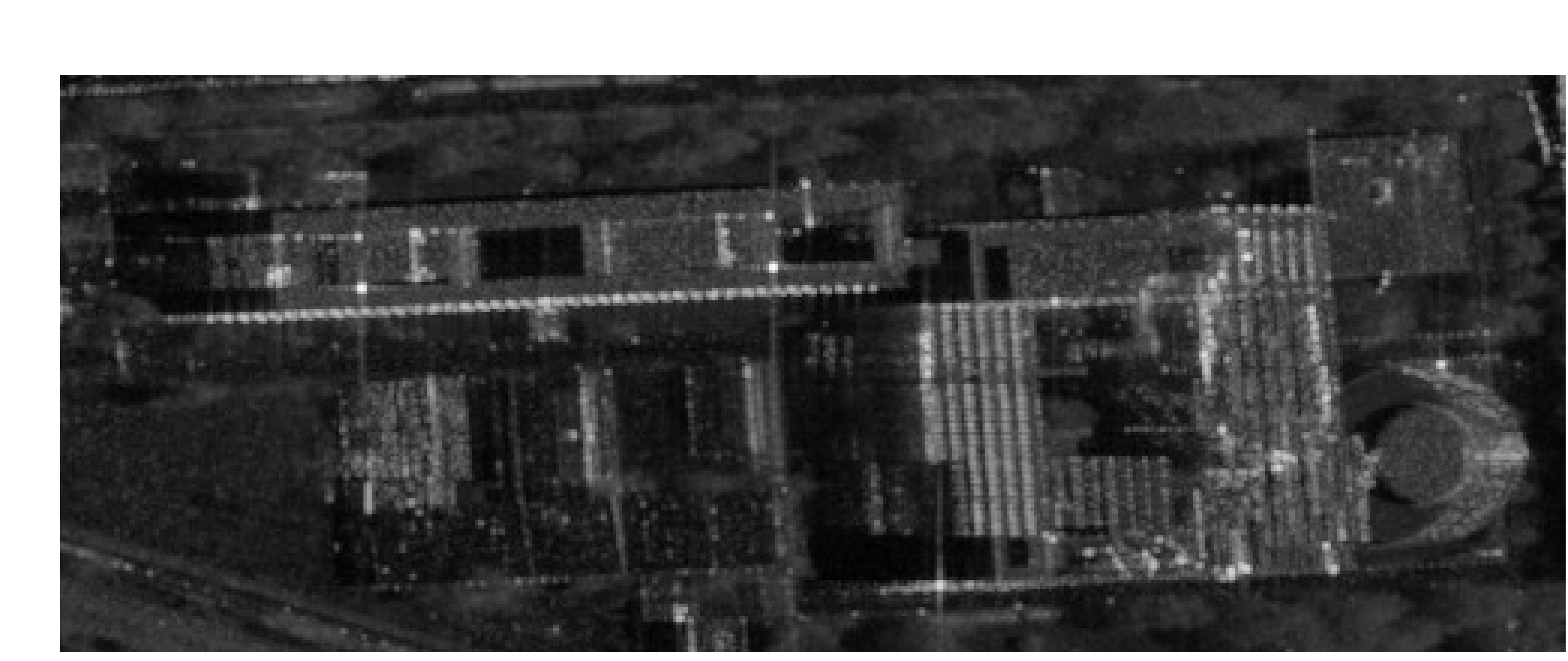}}
  \hfil
  \subfloat[]{\includegraphics[width=0.5\textwidth]{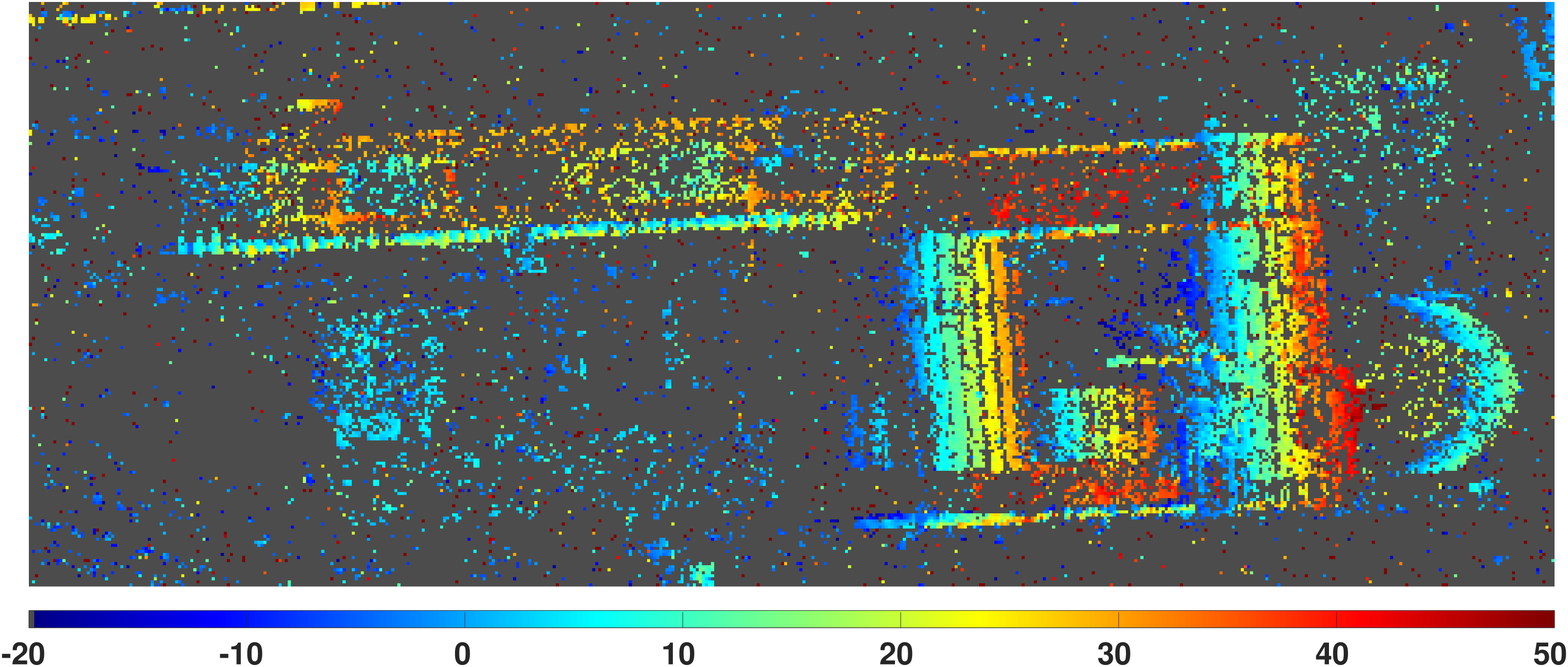}}
  \subfloat[]{\includegraphics[width=0.5\textwidth]{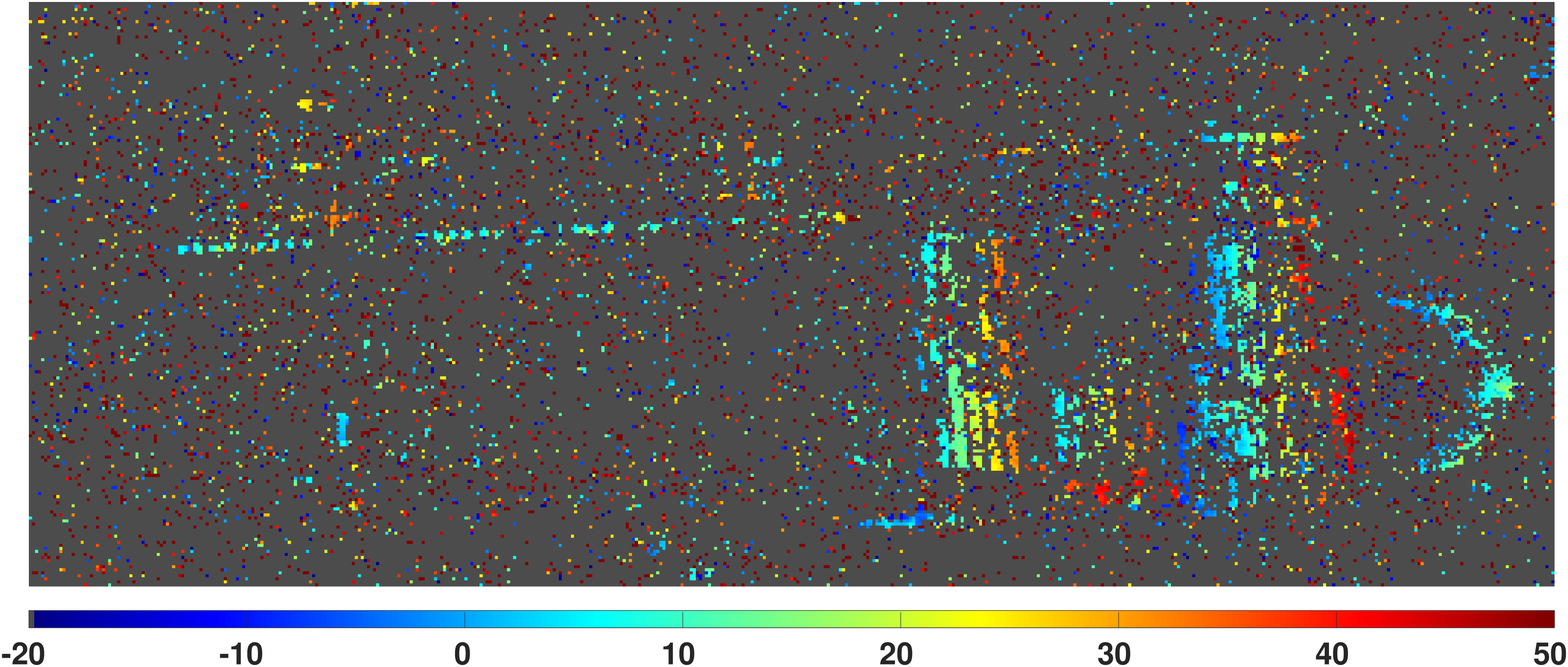}}
  \hfil
  \subfloat[]{\includegraphics[width=0.5\textwidth]{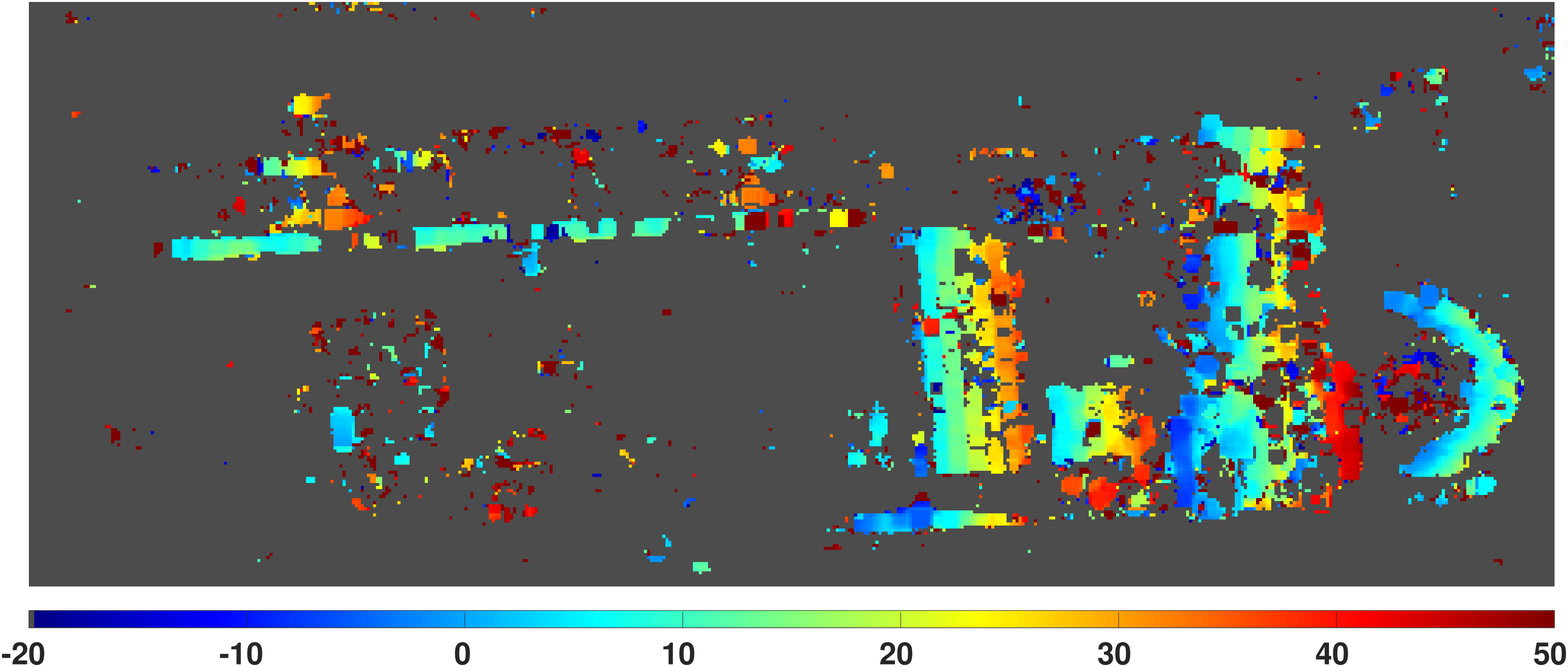}}
  \subfloat[]{\includegraphics[width=0.5\textwidth]{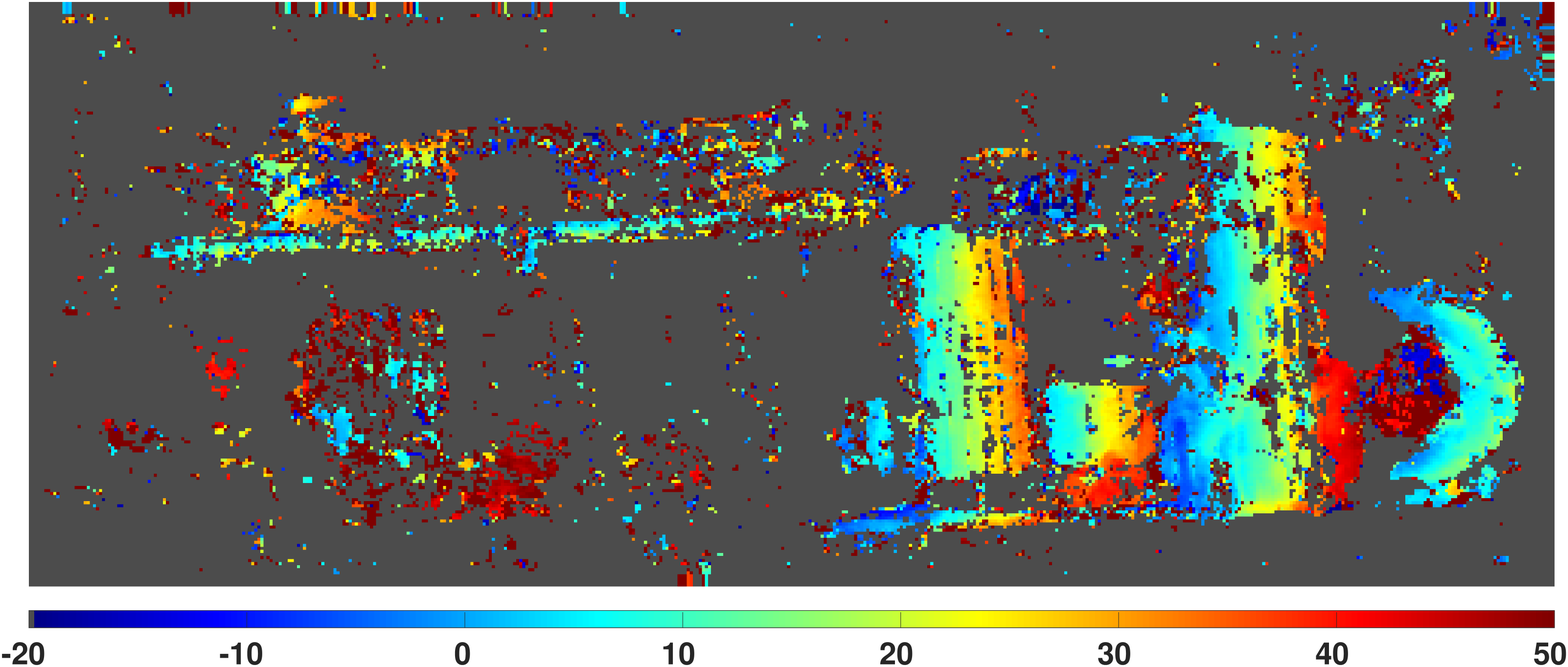}}
  \caption{Test building: DB Headquarters in Munich. (a) Optical image (Copyright Google). (b) Mean Amplitude. (c) Elevation estimated by SL1MMER with 64 images. (d) Elevation estimated by SL1MMER with 7 images. (e) Elevation estimated by Boxcar + SL1MMER with 7 images. (f) Elevation estimated by proposed method with 7 images.}
\label{fig:experiments_realData_DB}
\end{figure*}

\begin{figure}[!hb]
  \centering
\includegraphics[width=0.49\textwidth]{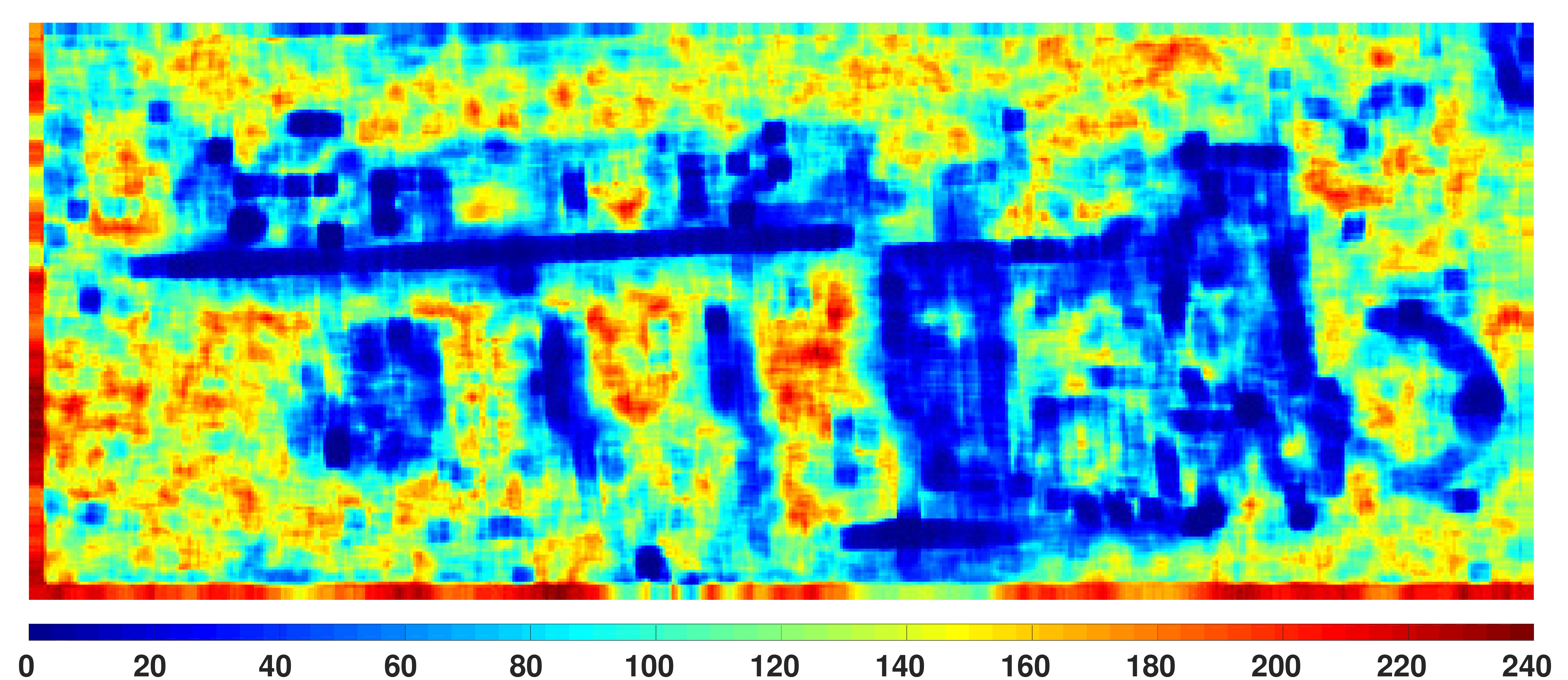}
\caption{Averaged equivalent number of looks by the non-local filter for the test site.}
\label{fig:experiments_realData_DB_numLooks}
\end{figure}

\begin{figure*}
  \centering
  \subfloat[]{\includegraphics[width=0.5\textwidth]{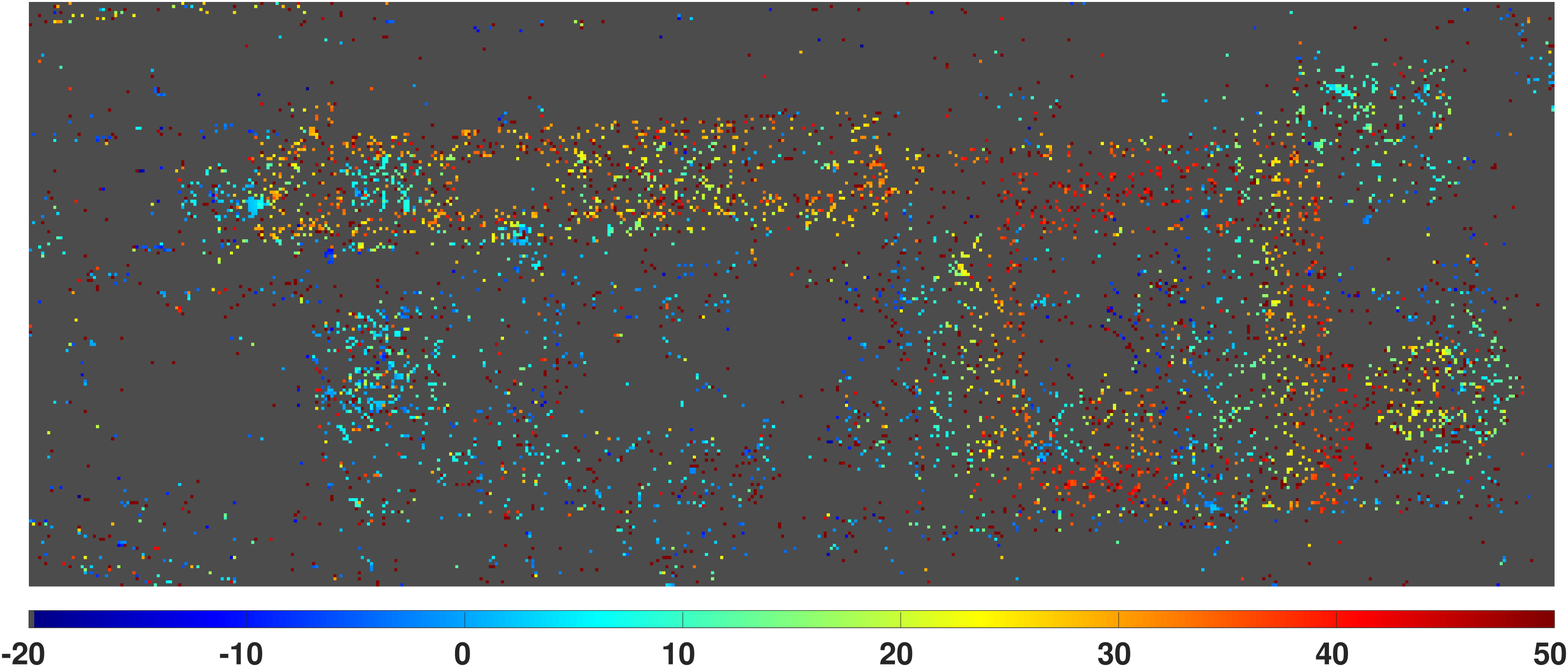}}
  \subfloat[]{\includegraphics[width=0.5\textwidth]{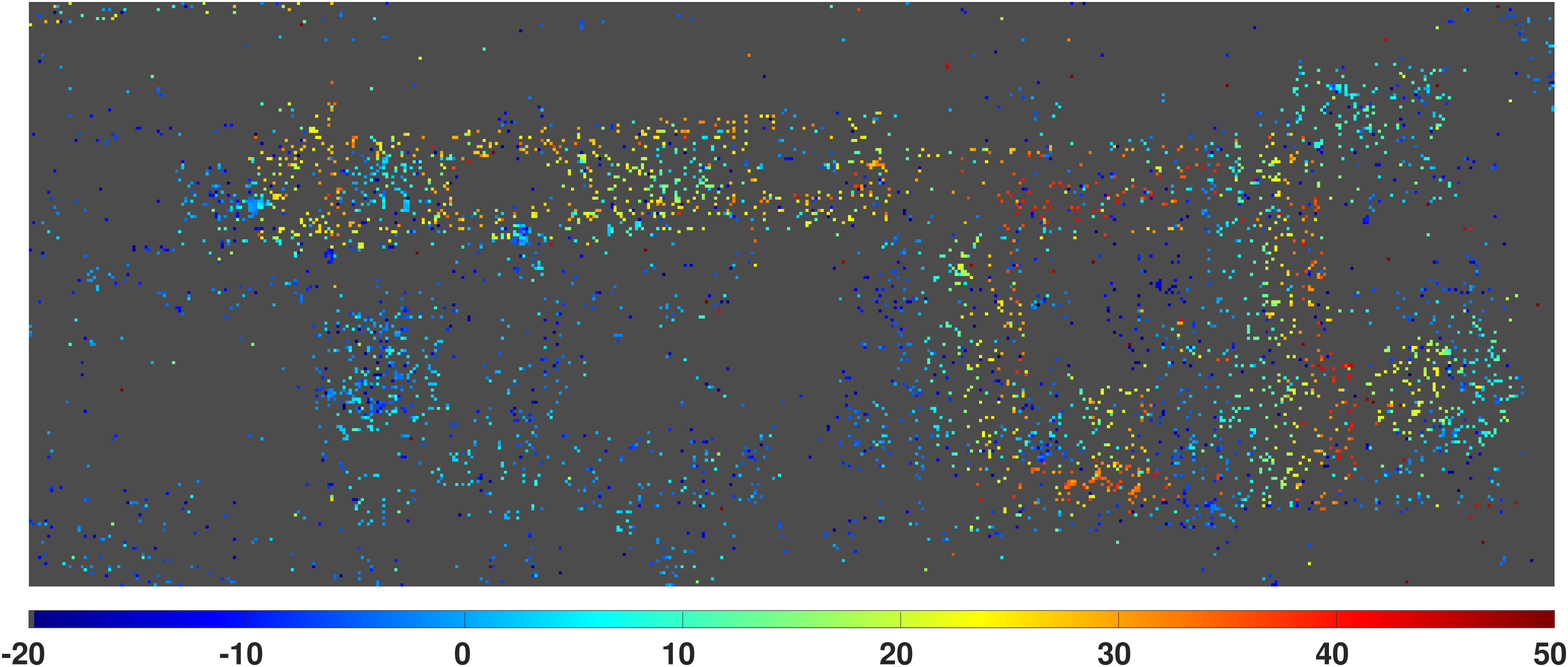}}
  \hfil
  \subfloat[]{\includegraphics[width=0.5\textwidth]{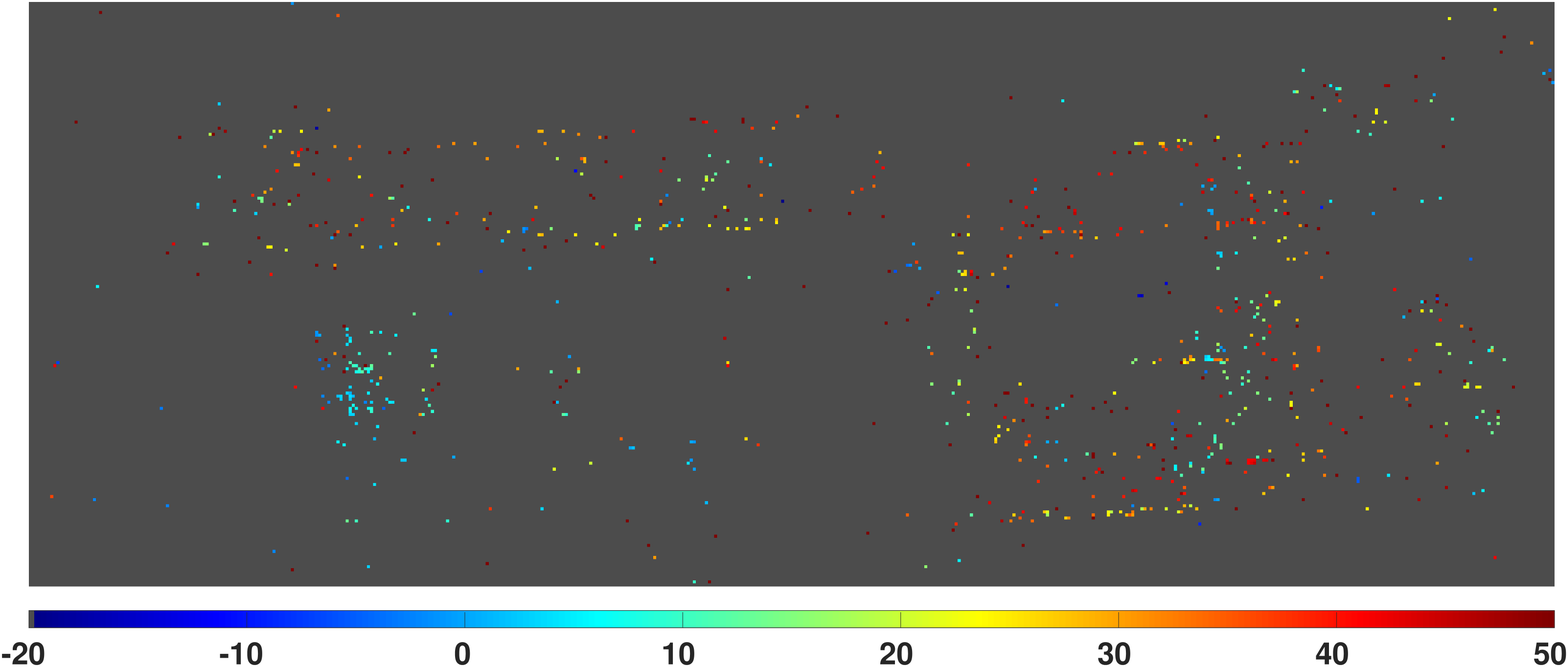}}
  \subfloat[]{\includegraphics[width=0.5\textwidth]{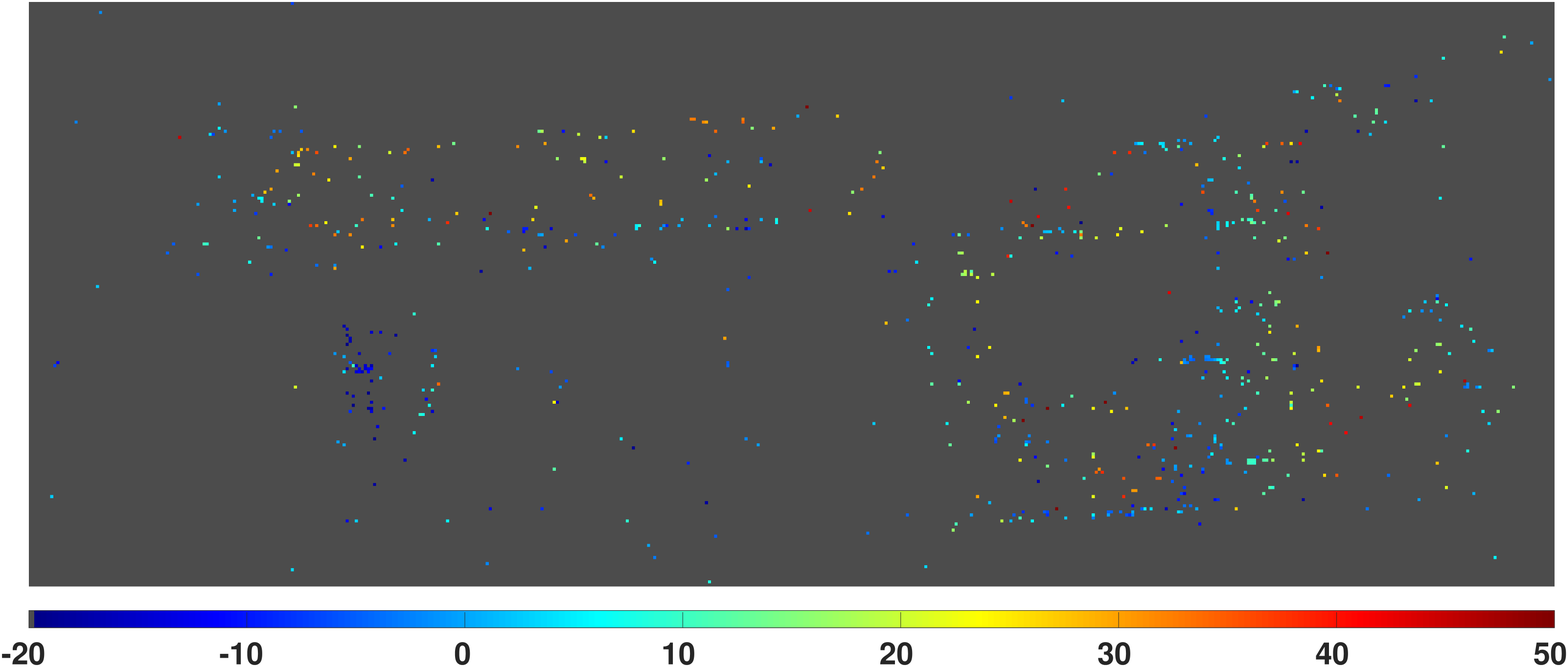}}
  \hfil
  \subfloat[]{\includegraphics[width=0.5\textwidth]{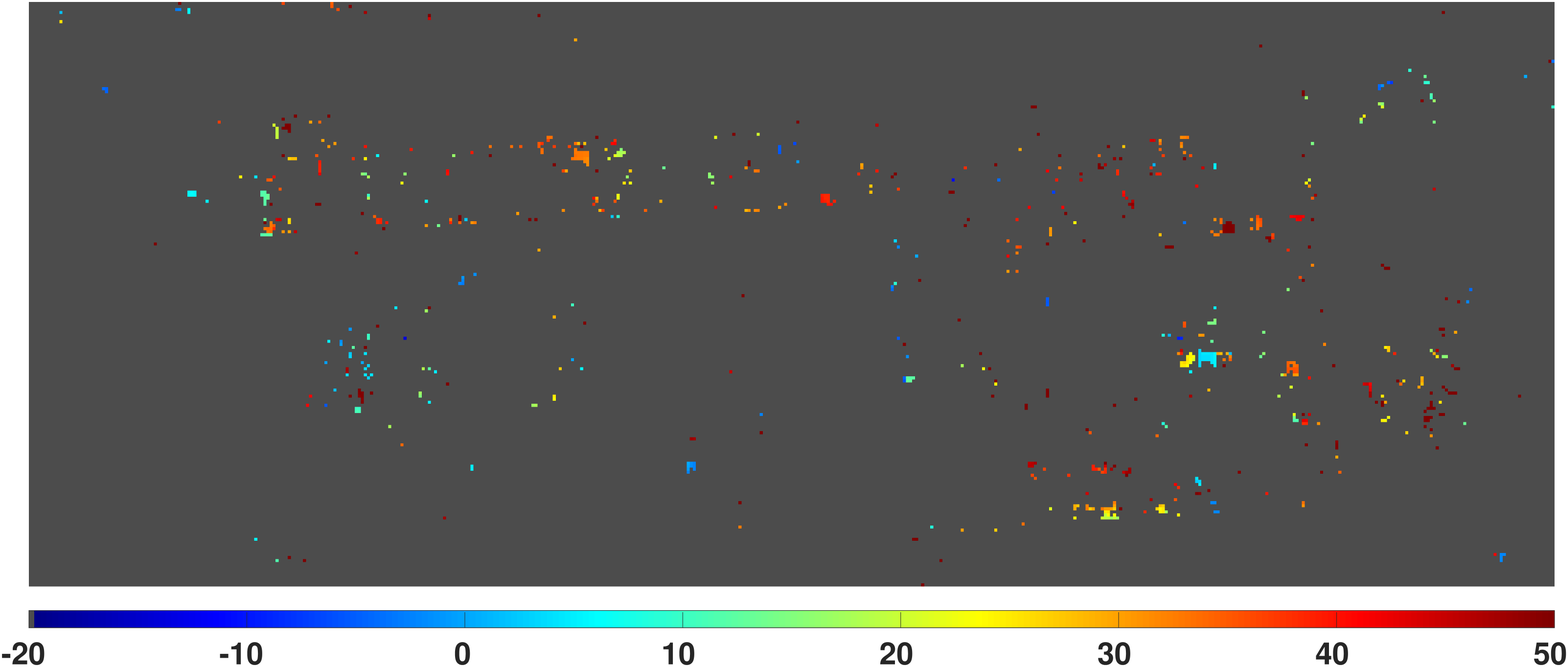}}
  \subfloat[]{\includegraphics[width=0.5\textwidth]{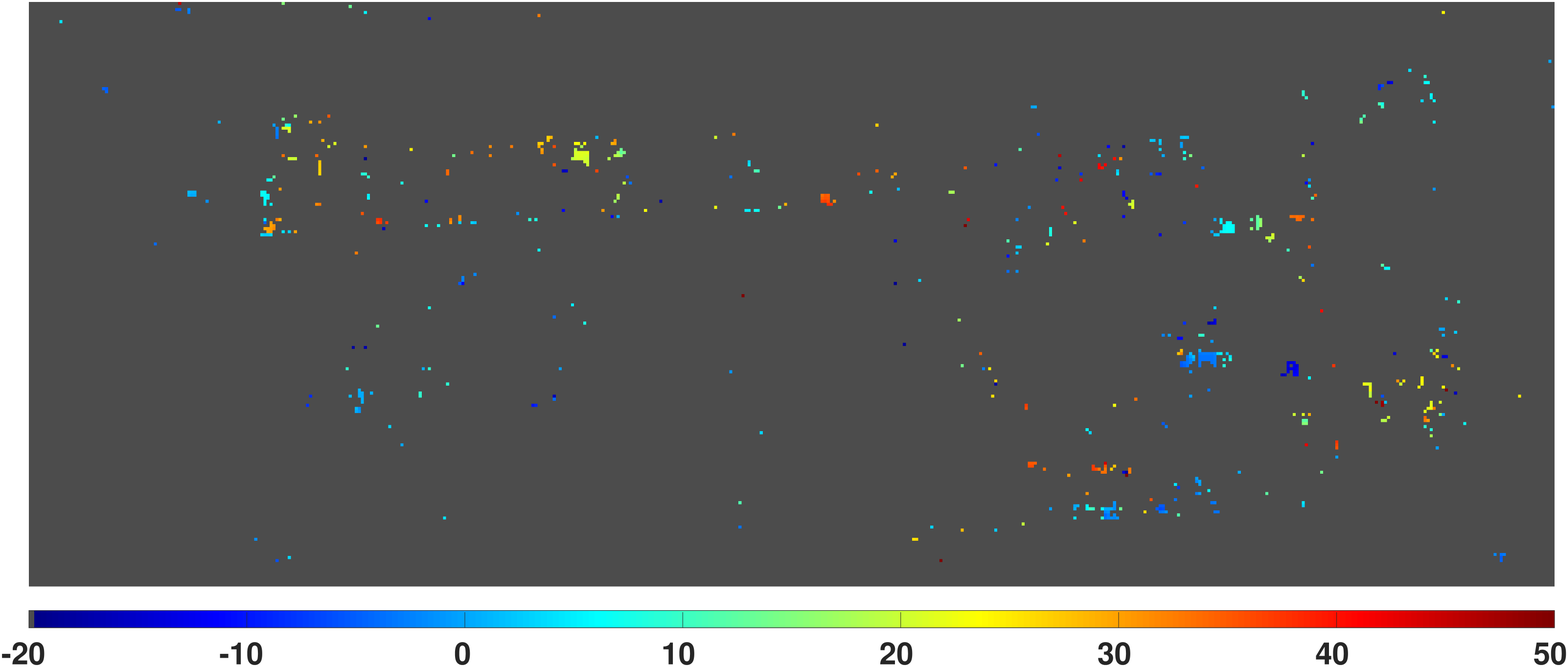}}
  \hfil
  \subfloat[]{\includegraphics[width=0.5\textwidth]{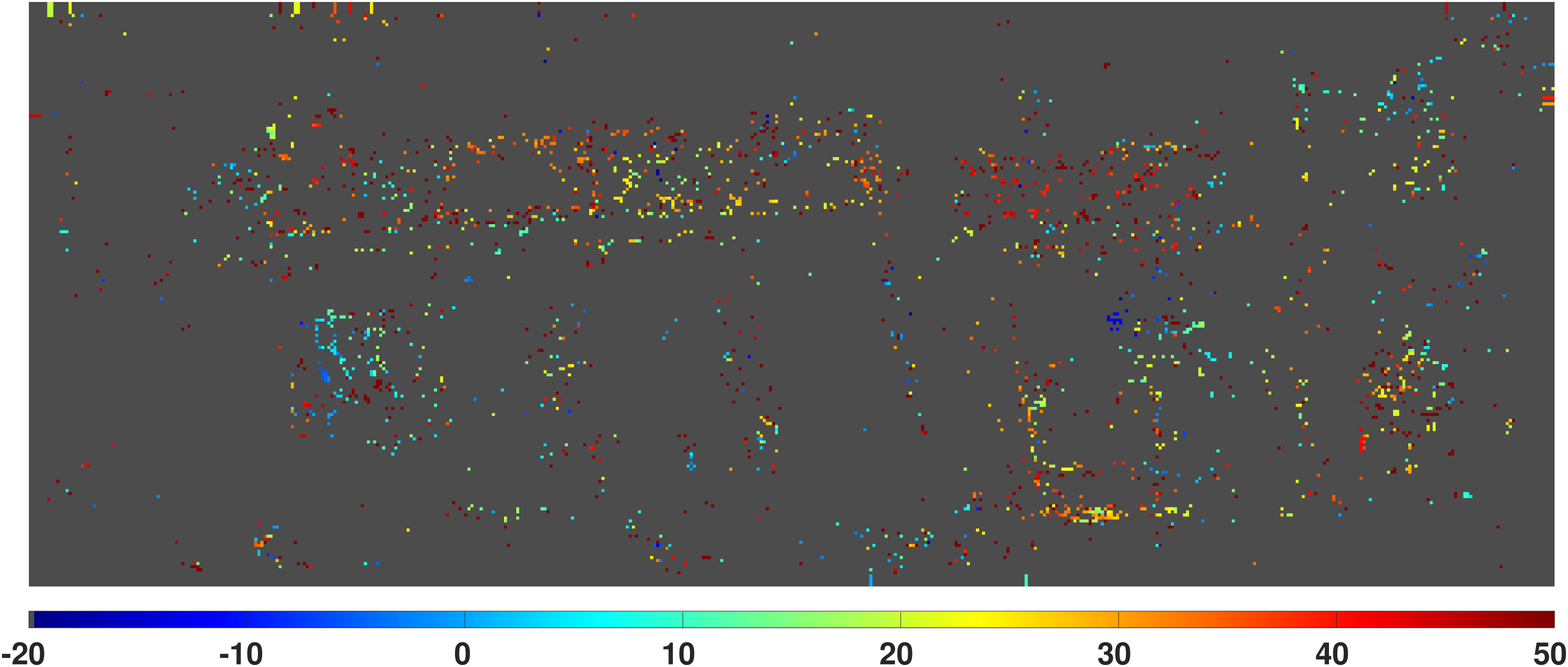}}
  \subfloat[]{\includegraphics[width=0.5\textwidth]{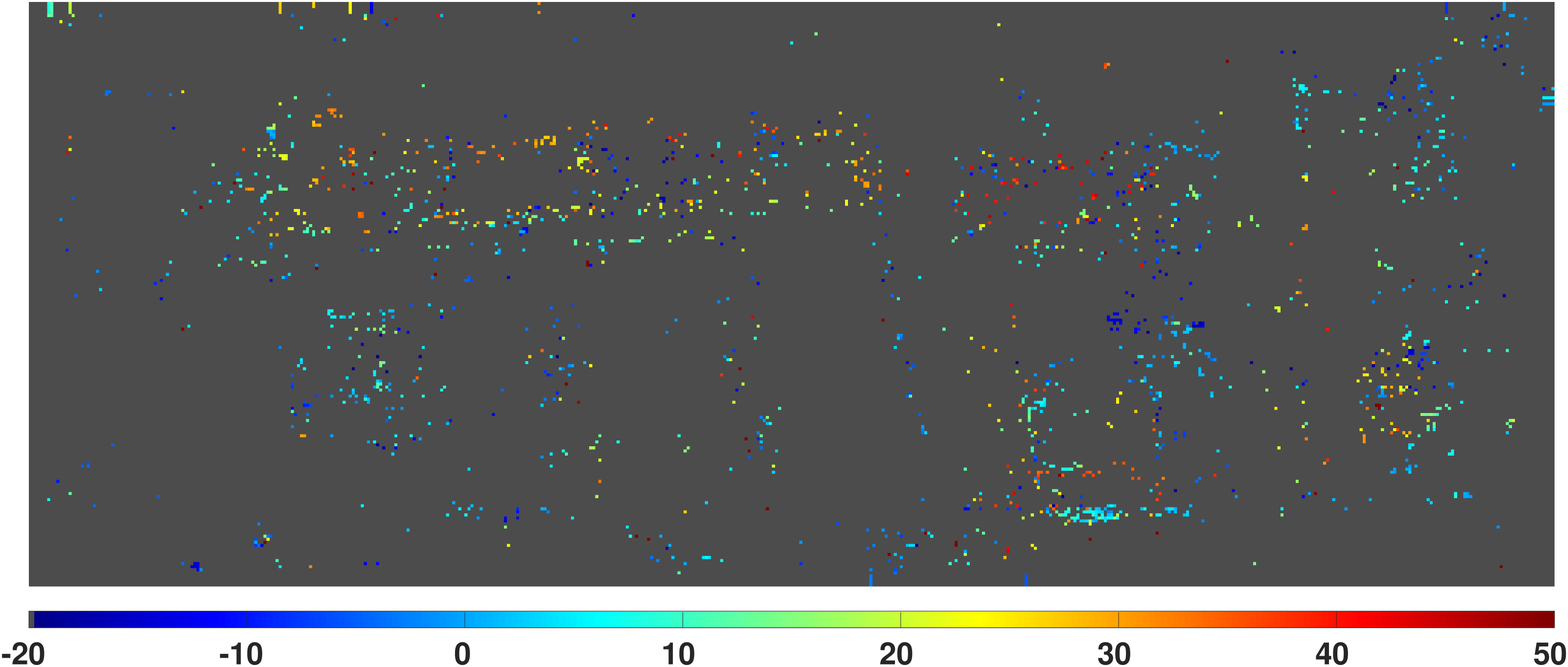}}
  \caption{Test building: DB Headquarter in Munich. Elevation estimates of the separated double scatterers. (a) Top layer (64 images + TomoSAR), mostly caused by returns from building facade. (b) Ground layer (64 images + TomoSAR), mostly caused by returns from ground structures. (c) Top layer (14 images + TomoSAR)  (d) Ground layer (14 images + TomoSAR)  (e) Top layer (14 images + Boxcar-TomoSAR)  (f) Ground layer (14 images + Boxcar-TomoSAR) (g) Top layer (14 images + NLCS-TomoSAR)  (h) Ground layer (14 images + NLCS-TomoSAR)}
  \label{fig:experiments_realData_DB_double}
\end{figure*}

The corresponding test area of the optical image and the mean amplitude of InSAR stack are shown in Figs. \ref{fig:experiments_realData_DB} (a) and (b), respectively. Fig. \ref{fig:experiments_realData_DB} (c) presents the elevation estimated by the SL1MMER approach with 64 images and Fig. \ref{fig:experiments_realData_DB} (d) presents the elevation estimated by the SL1MMER approach with 7 images. Note that the reconstructed result is not satisfactory with only 7 images. The estimated elevation exhibits strong noise due to the small number of images. And the successfully reconstructed elevation is significantly less than the reconstructed result with 64 images. As a comparison, we show the reconstructed result with a boxcar filter. Fig. \ref{fig:experiments_realData_DB} (e) is the elevation estimated by a boxcar filter with a window size of 5. It is clear that the loss of resolution is dramatic compared to the original TomoSAR. Boxcar filtering blurs edges and small structures present in the images. As can be seen in Fig. \ref{fig:experiments_realData_DB} (e), the proposed method can obtain an extraordinarily good estimated result with only 7 images. In contrast with the boxcar filter, the building structures are retrieved by our method, both in terms of shapes and elevations, without notable resolution loss.

Fig. \ref{fig:experiments_realData_DB_numLooks} shows the averaged equivalent number of looks by a non-local filter for the test site. From this figure, we can see that the elevation is estimated in a spatially adaptive manner. The number of looks at the buildings is quite lower than in the homogeneous area, which indicates that pixels chosen by a non-local filter should have similar properties, such as similar elevation, reflectivity, scattering characteristics, and so on.

In \cite{bib:Zhu2012b}, it was shown that a $90\%$ detection rate of two scatterers with a distance of $\rho_s$ can be achieved, while 17 acquisitions are needed when the amplitude of reflectivity of one scatterer is twice of another scatterer. In the real scenario, this is most often the case when one scatterer sits on the facade or roof and another is on the ground.
Fig. \ref{fig:experiments_realData_DB_double} presents elevation estimates of separated double scatterers. Figs. \ref{fig:experiments_realData_DB_double} (a) (c) (e) (g) show the top layer of scatterers reconstructed by TomoSAR with 64 images, TomoSAR with 14 images, Boxcar-TomoSAR with 14 images, and NLCS-TomoSAR with 14 images, respectively. Figs. \ref{fig:experiments_realData_DB_double} (b) (d) (f) (h) shows the ground layer of scatterers for the four cases. It is clear that TomoSAR with 64 images in Figs. \ref{fig:experiments_realData_DB_double} (a) (b) is proficient at reconstructing double scatterers, i.e., a top layer mostly caused by reflections from the facade of the building and a ground layer caused by reflections from lower buildings and ground infrastructures. As can be seen, keeping a similar SNR, TomoSAR with 14 images produces a very low detection rate for double scatterers in Figs. \ref{fig:experiments_realData_DB_double} (c) (d). After applying the boxcar filter, the detection rate increases a little bit, but the resolution decreases a lot. In contrast, the number of double scatterers detected by NLCS approach with small stacks is comparable to TomoSAR with large stacks, and its resolution loss is not obvious.

\begin{figure}
\centering
\includegraphics[width=0.45\textwidth]{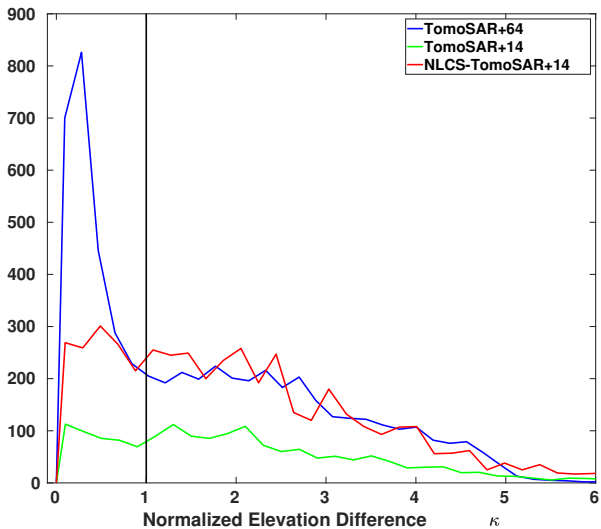}
\caption{Histogram of double scatterers' elevation differences using TomoSAR with 64 images (blue), TomoSAR with 14 images (green) and NLCS-TomoSAR with 14 images (red).}
\label{fig:histogram_double}
\end{figure}

The histogram of the double scatterers’ elevation differences using TomoSAR with 64 images (blue), TomoSAR with 14 images (green) and our NLCS-TomoSAR with 14 images (red) is shown in Fig. (10). The normalized distance is defined as
\begin{equation}
\kappa = \dfrac{s}{\rho_s}
\end{equation}
Note that the compressing-based SL1MMER reconstruction with the large stack has impressive SR capability, i.e., many of the double scatterers with $\kappa < 1$ are detected. However, when the number of interferograms decreases, the performance of double scatterer detection decreases accordingly. In contrast, the proposed NLCS-TomoSAR method with only 14 images obtains the same result as the large stack reconstruction, at least in the non-SR regime ($\kappa >1$). In the SR region, it is still much better than the standard method but falls short of the 64-stack reconstruction. We assume that this is caused by the averaging effect of NL filtering. Since target responses of different amplitude and different sub-pixel positions are averaged, the resulting amplitudes may be slightly compromised. The baseline-dependence of amplitude, however, is an important indicator of double scatterers.

\section{Conclusion}
In this work, we propose a novel framework for TomoSAR with a minimum number of acquisitions in order to obtain a fast and accurate estimation of elevation without any a priori knowledge. We evaluated the performance of the proposed NLCS algorithm with simulated and real data. Experiments using the simulated data illustrates that the proposed method can give excellent height estimation for different SNR without notable resolution distortion, in comparison to state of the art methods such as SL1MMER. Moreover, using only seven SAR images over the test site Munich, it is practically demonstrated that NLCS can achieve an accurate height estimation while preserving detailed structures. Furthermore, due to the increased SNR, a remarkable layover separation capability of NLCS can be observed.


\vfill

\begin{IEEEbiography}[{\includegraphics[width=1in,height=1.5in,clip,keepaspectratio]{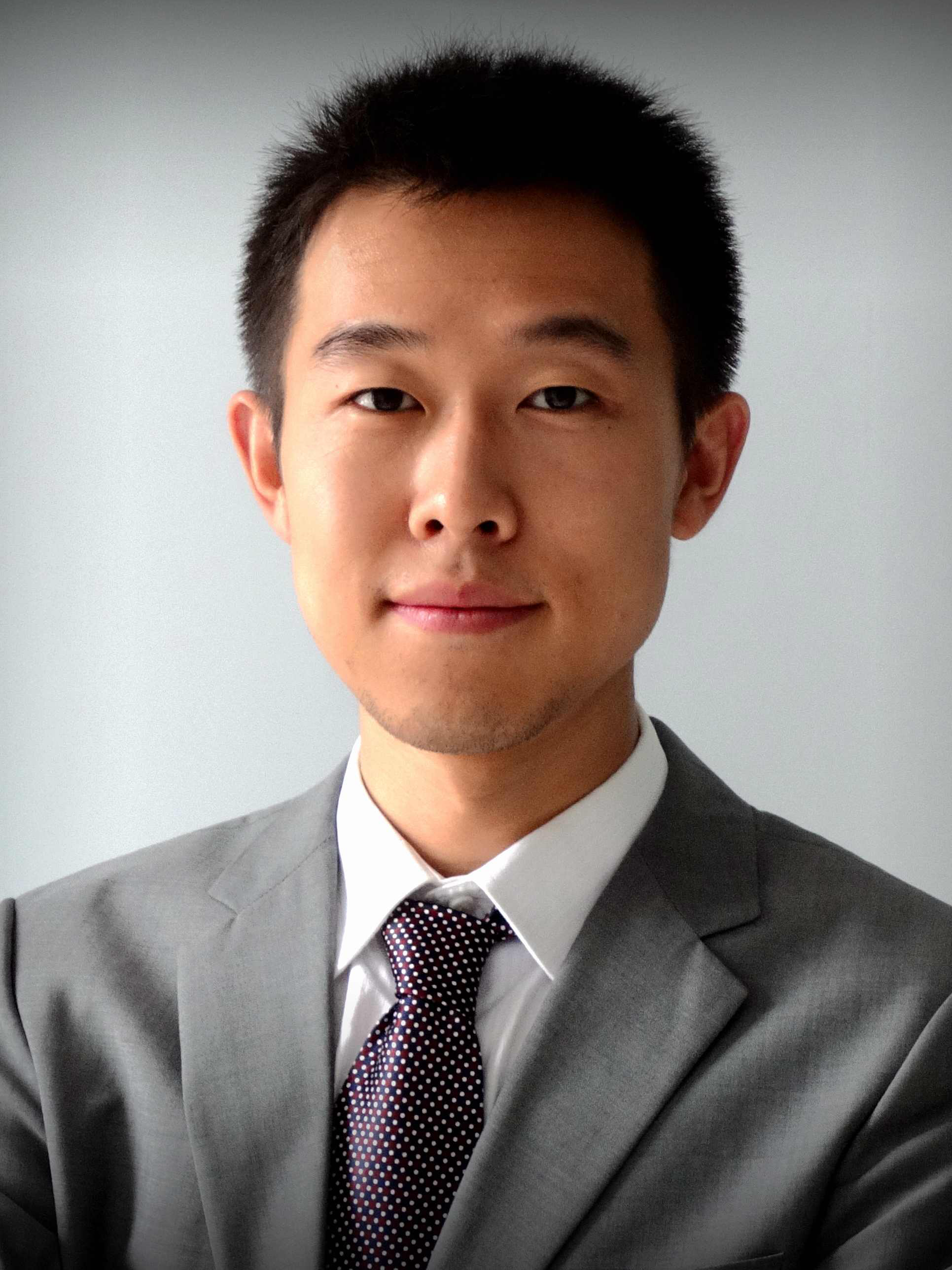}}]{Yilei Shi}
(M'18) received his Diploma degree in Mechanical Engineering from Technical University of Munich (TUM), Germany, in 2010. He is currently a research associate with the Chair of Remote Sensing Technology, Technical University of Munich.

His research interests include fast solver and parallel computing for large-scale problems, advanced methods on SAR and InSAR processing, machine learning and deep learning for variety data sources, such as SAR, optical images, medical images and so on; PDE related numerical modeling and computing.
\end{IEEEbiography}

\begin{IEEEbiography}[{\includegraphics[width=1in,height=1.5in,clip,keepaspectratio]{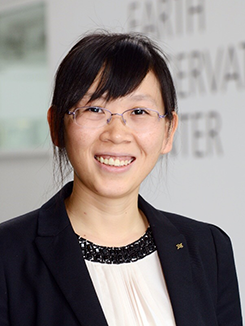}}]{Xiao Xiang Zhu}
(S'10--M'12--SM'14) received the Master (M.Sc.) degree, her doctor of engineering (Dr.-Ing.) degree and her “Habilitation” in the field of signal processing from Technical University of Munich (TUM), Munich, Germany, in 2008, 2011 and 2013, respectively.

She is currently the Professor for Signal Processing in Earth Observation (www.sipeo.bgu.tum.de) at Technical University of Munich (TUM) and German Aerospace Center (DLR); the head of the department of EO Data Science at DLR; and the head of the Helmholtz Young Investigator Group ”SiPEO” at DLR and TUM. Prof. Zhu was a guest scientist or visiting professor at the Italian National Research Council (CNR-IREA), Naples, Italy, Fudan University, Shanghai, China, the University  of Tokyo, Tokyo, Japan and University of California, Los Angeles, United States in 2009, 2014, 2015 and 2016, respectively. Her main research interests are
remote sensing and earth observation, signal processing, machine learning and data science, with a special application focus on global urban mapping.

Dr. Zhu is a member of young academy (Junge Akademie/Junges Kolleg) at the Berlin-Brandenburg Academy of Sciences and Humanities and the German National  Academy of Sciences Leopoldina and the Bavarian Academy of Sciences and Humanities. She is an associate Editor of IEEE Transactions on Geoscience and Remote Sensing.
\end{IEEEbiography}

\begin{IEEEbiography}[{\includegraphics[width=1in,height=1.5in,clip,keepaspectratio]{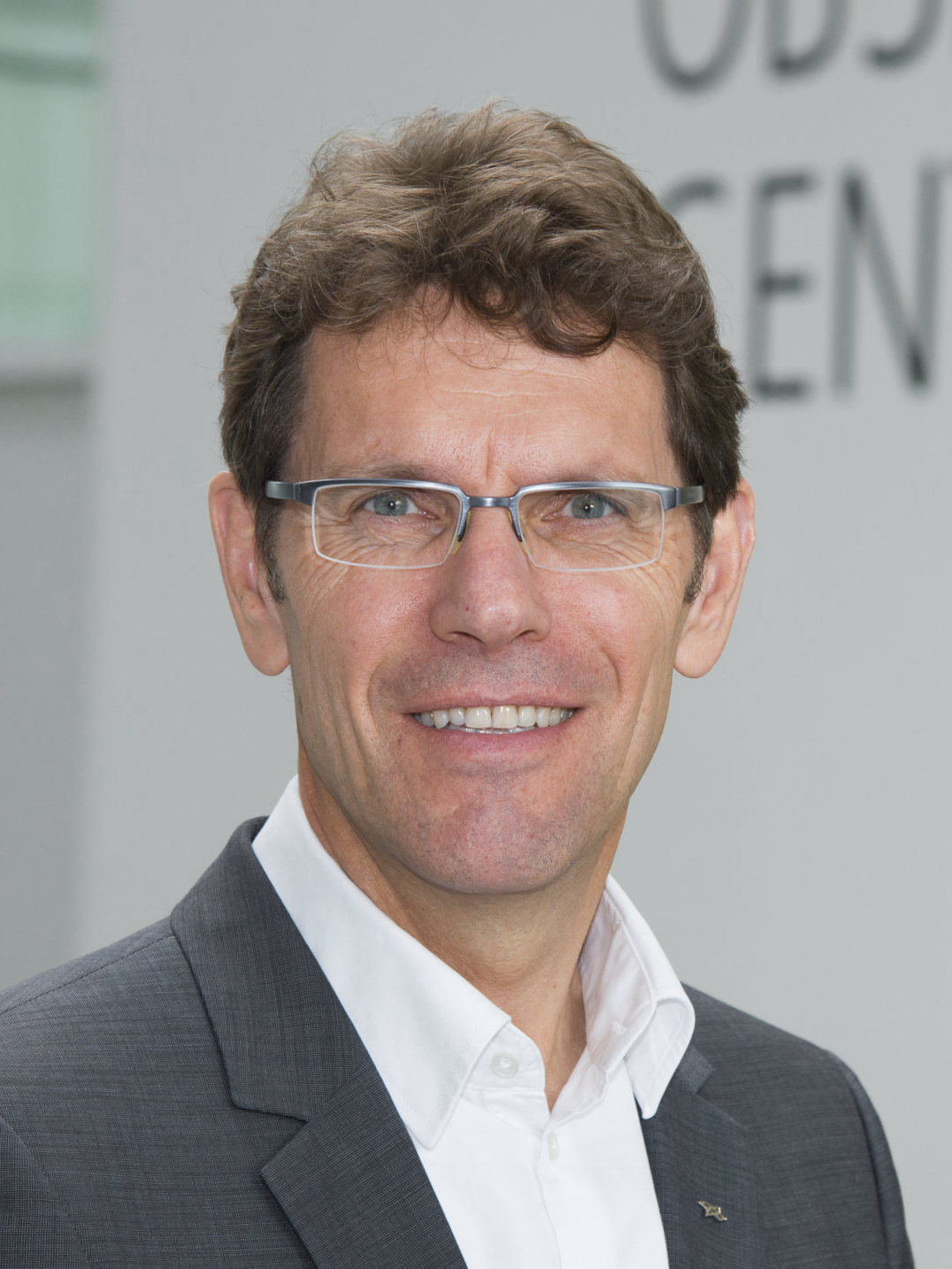}}]{Richard Bamler}
(M'95--SM'00--F'05) received his Diploma degree in Electrical Engineering, his Doctorate in Engineering, and his “Habilitation” in the field of signal and systems theory in 1980, 1986, and 1988, respectively, from the Technical University of Munich, Germany.

He worked at the university from 1981 to 1989 on optical signal processing, holography, wave propagation, and tomography. He joined the German Aerospace Center (DLR), Oberpfaffenhofen, in 1989, where he is currently the Director of the Remote Sensing Technology Institute.

In early 1994, Richard Bamler was a visiting scientist at Jet Propulsion Laboratory (JPL) in preparation of the SIC-C/X-SAR missions, and in 1996 he was guest professor at the University of Innsbruck. Since 2003 he has held a full professorship in remote sensing technology at the Technical University of Munich as a double appointment with his DLR position. His teaching activities include university lectures and courses on signal processing, estimation theory, and SAR. Since he joined DLR Richard Bamler, his team, and his institute have been working on SAR and optical remote sensing, image analysis and understanding, stereo reconstruction, computer vision, ocean color, passive and active atmospheric sounding, and laboratory spectrometry. They were and are responsible for the development of the operational processors for SIR-C/X-SAR, SRTM, TerraSAR-X, TanDEM-X, Tandem-L, ERS-2/GOME, ENVISAT/SCIAMACHY, MetOp/GOME-2, Sentinel-5P, Sentinel-4, DESIS, EnMAP, etc.

Richard Bamler’s research interests are in algorithms for optimum information extraction from remote sensing data with emphasis on SAR. This involves new estimation algorithms, like sparse reconstruction, compressive sensing and deep learning.
\end{IEEEbiography}

\end{document}